\documentclass{article}

\usepackage[main, final]{neurips_2026}
\usepackage[utf8]{inputenc} % allow utf-8 input
\usepackage[T1]{fontenc}    % use 8-bit T1 fonts
\usepackage{hyperref}       % hyperlinks
\usepackage{url}            % simple URL typesetting
\usepackage{booktabs}       % professional-quality tables
\usepackage{amsfonts}       % blackboard math symbols
\usepackage{nicefrac}       % compact symbols for 1/2, etc.
\usepackage{microtype}      % microtypography
\usepackage{xcolor}         % colors
\usepackage{enumitem}%
\usepackage{amsmath}%
\usepackage{amssymb}%
\usepackage{bm}%
\usepackage{multirow}
\usepackage{graphicx}
\usepackage{wrapfig}
\newcommand{\sd}[1]{$_{\pm #1}$}%
\title{Rethinking Contrastive Loss in CLIP Post-training: A Complementary Framework with Frozen Text Encoder}

\author{%
  Zidan Wang\textsuperscript{1} \quad 
  Yaqian Li\textsuperscript{2,$*$} \quad
  Xiaokai Zhang\textsuperscript{1} \quad 
  Kaiwen Long\textsuperscript{2} \quad
  Kun He\textsuperscript{1} \quad
  Hanpeng Liu\textsuperscript{1,2,$*$,$\dagger$} \\
  \textsuperscript{1}Huazhong University of Science and Technology \quad
  \textsuperscript{2}Li Auto Inc. \\
  \texttt{liyaqian@lixiang.com} \quad \texttt{hanpengliu@hust.edu.cn}
}

\begin{document}

\maketitle
% author notes (set manually so that the marks in the author line do not overlap)
\begingroup
\renewcommand{\thefootnote}{\fnsymbol{footnote}}
\footnotetext[1]{Corresponding authors.}
\footnotetext[2]{Work done during a research internship at Li Auto Inc.}
\endgroup

\begin{abstract}
CLIP serves as a foundational vision-language model and the de facto
vision encoder for downstream VLMs such as LLaVA. Post-training
offers a lightweight route to refine CLIP, but recent work
argues that the standard contrastive loss is unsuitable for
post-training due to catastrophic forgetting under small batches,
motivating designs that abandon the contrastive objective in favor
of distillation. We revisit this premise and find that, for the
InfoNCE objective, the reported forgetting is driven primarily not by
insufficient negatives but by an inappropriate magnitude of the
contrastive temperature $\tau$: with $\tau$ set sufficiently small,
contrastive post-training improves rather than degrades the
pretrained CLIP, which we explain through the temperature dependence
of the InfoNCE gradient. Building on this finding, we propose
\textbf{ComCLIP}, a lightweight single-epoch post-training recipe
that freezes CLIP's text encoder---so the refined vision encoder is
a drop-in replacement with unchanged architecture and inference
cost---and trains the vision encoder with a properly-tempered
contrastive loss, an MSE anchoring loss against the original CLIP,
and a relational distillation loss from DINOv2. Over multiple seeds,
ComCLIP matches the self-distillation baseline CLIP-Refine on
zero-shot classification while significantly improving the
transferability of visual features, measured by linear probing
($48.99$ vs.\ $42.28$ on ViT-B/16), and on ViT-L/14 it also improves MMVP over CLIP-Refine
($24.20$ vs.\ $19.01$); CLIP-Refine remains
stronger on image-text retrieval. Used as a drop-in vision encoder
for LLaVA-1.5-7B without re-aligning the projector or LLM, ComCLIP
yields no net change across $8$ VLM benchmarks, i.e., the refinement
does not break downstream compatibility. Code and models are available at
\url{https://github.com/showstarpro/ComCLIP.git}.
\end{abstract}

\section{Introduction}
\label{sec:intro}

Vision-language foundation models, particularly CLIP~\citep{radford2021clip}, 
serve as the de facto vision encoder for modern vision-language models 
(VLMs) such as LLaVA~\citep{liu2023llava}. Despite its success, CLIP 
exhibits residual issues including the modality 
gap~\citep{liang2022mind} and systematic visual blindness on 
fine-grained pairs~\citep{tong2024mmvp}, which propagate into 
downstream VLMs. Re-training CLIP from 
scratch~\citep{zhai2023siglip,sun2023evaclip} is prohibitively 
expensive and breaks compatibility with the existing CLIP-based 
ecosystem. \emph{Post-training}---refining a pretrained CLIP on a 
small dataset with limited compute---offers a pragmatic alternative.
We target a developer who upgrades the vision encoder of a CLIP-based
VLM without retraining its projector or LLM: the refined encoder must
stay in the original CLIP image-text space, and the visual features
the VLM consumes should become more transferable.

Existing CLIP post-training methods fall into two categories: 
self-distillation methods such as CLIP-Refine~\citep{cliprefine2024} 
that use image-text soft labels, and external-teacher distillation 
methods such as KUEA~\citep{uea2024} that align the geometric 
structure of CLIP's visual representations with 
DINOv2's~\citep{oquab2023dinov2} via kernel-based methods. Notably, 
both lines deliberately \emph{avoid} the standard image-text 
contrastive loss---the very objective that produced CLIP. 
CLIP-Refine, for instance, argues that contrastive post-training is 
fundamentally limited by the small batch sizes affordable in this 
regime, which provide insufficient negative samples and lead to 
catastrophic forgetting; they report that even scaling the batch 
size up to $2{,}048$---near the practical ceiling for 
post-training---still fails to recover the zero-shot performance of 
the original CLIP. This pessimistic conclusion about contrastive 
loss has shaped subsequent designs that rely on distillation rather 
than direct contrastive learning.

\begin{wrapfigure}{r}{0.52\textwidth}
    \centering
    \vspace{-12pt} 
    \includegraphics[width=0.5\textwidth]{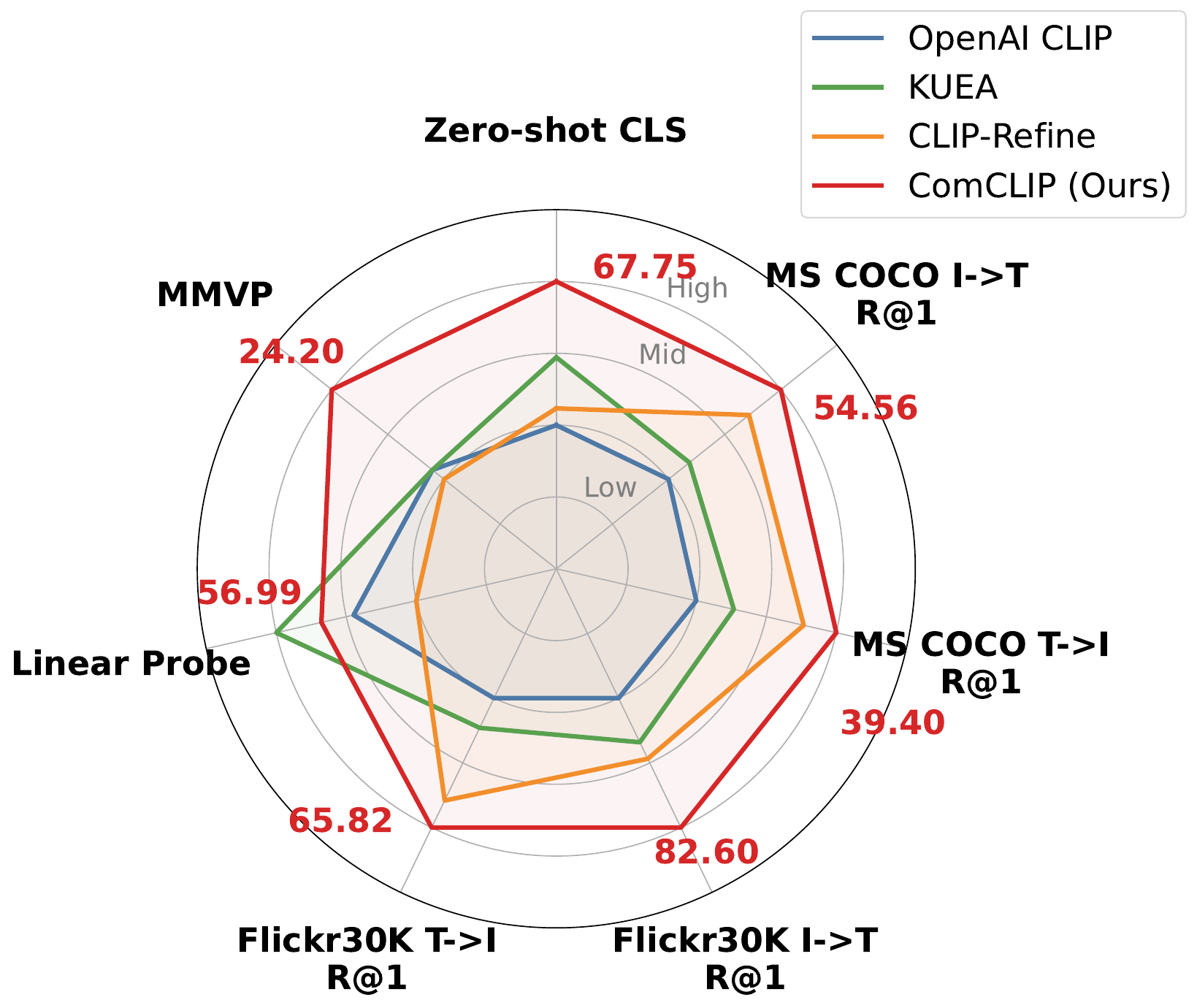}
    \caption{Multi-dimensional comparison on ViT-L/14 (values from 
    Table~\ref{tab::main_results}). Relative to the original OpenAI 
    CLIP, ComCLIP improves all four evaluation dimensions (zero-shot 
    classification, retrieval, linear probing, MMVP), whereas prior 
    post-training methods exhibit characteristic trade-offs. Each axis
    is min--max normalized per metric; ComCLIP's MMVP is the $3$-seed
    mean (Sec.~\ref{sec:exp-seeds}).}
    \label{fig:radar}
    \vspace{-10pt} 
\end{wrapfigure}

\textbf{Finding 1: Contrastive temperature is a major underappreciated 
factor in CLIP post-training.} We revisit this premise and find that, 
for the InfoNCE objective, the reported forgetting is driven 
primarily not by insufficient negatives in small batches but by an 
inappropriate magnitude of the contrastive temperature $\tau$. 
Directly inheriting the pretraining initialization $\tau \approx 0.07$ 
(i.e., $1/\tau \approx 14$) leads to performance degradation, whereas 
a substantially smaller $\tau = 0.01$ (i.e., $1/\tau = 100$) does not. 
The mechanism is visible in the InfoNCE gradient: the gradient on a 
negative pair decays as $\exp(-(s_{ii}-s_{ij})/\tau)$ in its margin to 
the positive. Because a pretrained CLIP already separates matched from 
mismatched pairs, a small $\tau$ makes the gradient on easy pairs 
vanish, so the pretrained geometry is left largely intact, and 
concentrates the update on the few hard negatives, which refines it 
(Sec.~\ref{sec:method-clip}). On ViT-B/16 (CC3M, $1$ epoch, batch 
size $1{,}024$), naive contrastive post-training with $\tau = 0.07$ 
drops $12$-dataset zero-shot accuracy from $61.82$ to $57.90$, 
reproducing prior reports; with $\tau = 0.01$, the same objective 
improves zero-shot accuracy to $62.90$ under the same compute budget.

\textbf{Finding 2: Different post-training objectives act on 
separable metric families.} While properly-tempered contrastive 
loss alone is beneficial, it leaves the transferability of visual 
features, measured by linear probing, essentially unchanged. 
Our loss-decoupling ablation shows that the objectives we consider 
act on largely separable groups of metrics: the contrastive loss 
drives zero-shot classification and retrieval, anchoring to the 
pretrained encoder preserves transferable visual features, and 
relational distillation from DINOv2 adds further 
linear-probe transferability. We do not claim this is a unique or 
exhaustive decomposition of post-training failure modes; it is the 
empirical basis for combining the objectives. 
Building on these findings, we propose \textbf{ComCLIP}, a 
lightweight single-epoch post-training recipe that freezes the text 
encoder---so the refined vision encoder is a drop-in replacement with 
unchanged architecture and inference cost---and trains the vision 
encoder under three losses that follow one principle, 
\emph{refine relative structure while anchoring absolute geometry}: 
a properly-tempered contrastive loss for alignment, an MSE 
anchoring loss against the original CLIP, and a relational 
distillation loss from DINOv2.

On ViT-B/16 and ViT-L/14 with one epoch on CC3M or COCO Caption, 
ComCLIP and the self-distillation baseline CLIP-Refine sit at 
different points of a trade-off. Over three seeds on ViT-B/16, the 
two are at parity on $12$-dataset zero-shot accuracy 
($63.00_{\pm0.13}$ vs.\ $62.87_{\pm0.11}$, not significant), 
CLIP-Refine is ahead on image-text retrieval, and ComCLIP is 
substantially ahead on linear probing ($48.99_{\pm0.35}$ vs.\ 
$42.28_{\pm0.62}$, $p<0.001$), an advantage that also holds when both 
methods use the same post-training data. On ViT-L/14, ComCLIP 
additionally improves MMVP over CLIP-Refine trained on the same data 
($24.20_{\pm0.86}$ vs.\ $19.01_{\pm1.86}$). Swapped into LLaVA-1.5-7B 
without any re-alignment, ComCLIP yields no net change across $8$ VLM 
benchmarks: the refinement preserves downstream compatibility rather 
than improving downstream accuracy.

\noindent
\textbf{Contributions.}
\textbf{(1)} We show that, for InfoNCE, the failure of contrastive 
post-training reported in prior work is driven primarily by the 
magnitude of $\tau$ rather than by batch size, isolate this effect 
from batch size and temperature learnability, and explain it through 
the temperature dependence of the InfoNCE gradient. 
\textbf{(2)} We propose \textbf{ComCLIP}, a lightweight 
single-epoch post-training recipe with a frozen text encoder that 
combines a properly-tempered contrastive loss with anchoring and 
relational distillation. 
\textbf{(3)} With multi-seed statistics, we show that ComCLIP 
significantly improves linear-probe transferability over CLIP-Refine 
at zero-shot parity on ViT-B/16 and ViT-L/14, improves MMVP over 
CLIP-Refine on ViT-L/14, and serves as a drop-in vision encoder for 
LLaVA-1.5-7B without degrading downstream performance.

\section{Related Work}
\label{sec:related}

\paragraph{CLIP and its limitations.}
CLIP~\citep{radford2021clip} pioneered web-scale contrastive
image-text pretraining and serves as the de facto vision encoder
for modern VLMs~\citep{liu2023llava}. While subsequent work
improved pretraining via better objectives
(SigLIP~\citep{zhai2023siglip}), rewritten or cleaner data
(LaCLIP~\citep{fan2023laclip}), more efficient training recipes
(CLIPA~\citep{li2023clipa}), and stronger
architectures (EVA-CLIP~\citep{sun2023evaclip}), CLIP still
exhibits the modality gap~\citep{liang2022mind} and visual
blindness on fine-grained pairs~\citep{tong2024mmvp}, both of
which propagate into downstream VLMs~\citep{li2024promptkd}.
These pretraining studies also document that the temperature
interacts with batch size and training recipe, which provides
context for our Finding~1; our setting differs in that the model
starts from a converged CLIP and sees only a small corpus for a
single epoch.

\paragraph{Freezing one tower.}
LiT~\citep{zhai2022lit} freezes a pretrained \emph{image} encoder and
tunes the text encoder to it. ComCLIP is the mirror image: the
\emph{text} encoder is frozen and the image encoder is refined, so
that the refined encoder remains compatible with every system built
on the original CLIP text space (e.g., LLaVA's projector).

\paragraph{CLIP post-training.}
Post-training refines a pretrained CLIP with limited compute.
\textbf{Self-distillation} methods such as
CLIP-Refine~\citep{cliprefine2024} use image-text soft labels and
explicitly abandon the contrastive objective, arguing it causes
catastrophic forgetting under small batches.
\textbf{External-teacher distillation} such as KUEA~\citep{uea2024}
aligns CLIP with DINOv2~\citep{oquab2023dinov2} via kernel-based
methods, but risks degrading the language-aligned contrastive
structure.
\textbf{Generative-model-based enhancement}---DIVA~\citep{diva2024},
un2CLIP~\citep{un2clip2024}, GenHancer~\citep{genhancer2024}---uses
feedback from diffusion or unCLIP-style generative models to improve
fine-grained visual perception, with large gains on MMVP. Their
published results show that this need not come at the cost of CLIP's
other capabilities: DIVA and GenHancer report essentially unchanged
zero-shot classification and retrieval, and un2CLIP reports
improved text-to-image retrieval with reduced ImageNet accuracy
(Appendix~\ref{app:generative}). These methods rely on a large
generative model during training and are therefore in a heavier
compute class than single-epoch contrastive post-training.
\textbf{Retrieval-augmented customization} such as
REACT~\citep{liu2023react} retrieves task-relevant image-text pairs
from a web-scale corpus to adapt CLIP to a target domain; it
addresses domain customization rather than general-purpose
refinement and is not directly comparable.

\paragraph{Our position.}
ComCLIP revisits the role of the contrastive loss within
lightweight post-training at a fixed text encoder. It shows that
the InfoNCE objective is a viable post-training signal when its
temperature is set appropriately, and combines it with anchoring
and relational distillation~\citep{park2019rkd,tung2019sp,zheng2021ressl} from
DINOv2. We do not aim to maximize fine-grained perception, where
generative methods lead; our target is a single-epoch refinement
that improves the transferability of the visual features while
keeping the encoder a drop-in replacement.

\section{Method}
\label{sec:method}

\begin{figure}[t!]
\centering
\includegraphics[width=1.0\linewidth]{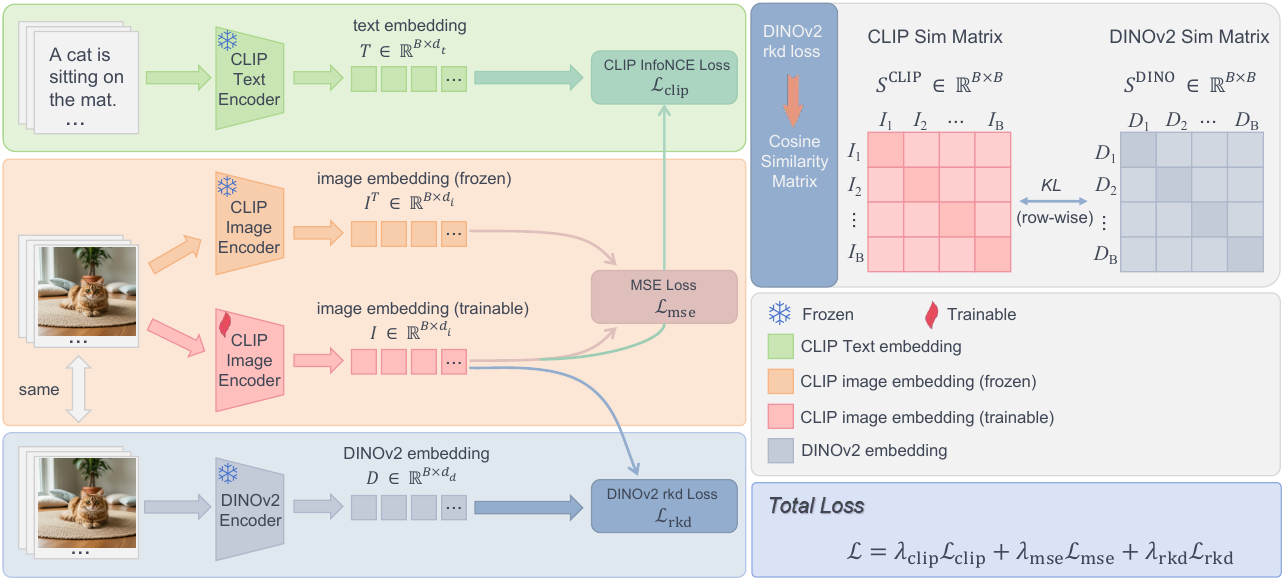}
\caption{\textbf{Overview of ComCLIP.} Three parallel branches 
operate jointly: (i) frozen CLIP text encoder for image-text 
alignment ($\mathcal{L}_{\text{clip}}$), (ii) trainable CLIP image 
encoder anchored to its frozen pretrained counterpart 
($\mathcal{L}_{\text{mse}}$), and (iii) frozen DINOv2 encoder 
providing relational supervision via batch-wise similarity 
matrices ($\mathcal{L}_{\text{rkd}}$). All three losses operate on 
the post-projection embedding.}
\label{fig:comclip}
\vspace{-1em}
\end{figure}

We propose \textbf{ComCLIP}, a lightweight post-training recipe 
that refines CLIP's vision encoder under three losses that act on 
separable metric families (Sec.~\ref{sec:intro}). The text encoder is frozen throughout to 
preserve drop-in compatibility with downstream VLMs. The 
framework is illustrated in Figure~\ref{fig:comclip}.

\subsection{Preliminaries}
\label{sec:method-prelim}

CLIP~\citep{radford2021clip} consists of an image encoder $f_v$ 
and a text encoder $f_t$, jointly trained on image-text pairs 
$(x, t)$ via the symmetric InfoNCE loss~\citep{oord2018cpc}:
\begin{equation}
    \mathcal{L}_{\text{NCE}}(\bm{u}, \bm{v}; \tau) = 
    -\frac{1}{B} \sum_{i=1}^{B} \log 
    \frac{\exp(\bm{u}_i^\top \bm{v}_i / \tau)}
    {\sum_{j=1}^{B} \exp(\bm{u}_i^\top \bm{v}_j / \tau)},
    \label{eq:infonce}
\end{equation}
where $\bm{u}_i, \bm{v}_i \in \mathbb{R}^d$ are $\ell_2$-normalized 
projected embeddings, $B$ is the batch size, and $\tau > 0$ is 
the contrastive temperature. The full CLIP objective is symmetric 
over images and texts. For each image $x$, we denote the 
post-projection $\ell_2$-normalized embedding produced by $f_v$ as 
$\bm{u}(x) \in \mathbb{R}^d$, which lives in the shared image-text 
contrastive space. Given a pretrained CLIP $(f_v^0, f_t^0)$ and a 
small post-training dataset $\mathcal{D}$, our goal is to refine 
$f_v$ while keeping $f_t^0$ frozen, with all ComCLIP losses 
operating on $\bm{u}$.

\subsection{Overview of ComCLIP}
\label{sec:method-overview}

ComCLIP trains $f_v$ under a weighted sum of three losses:
\begin{equation}
    \mathcal{L}_{\text{ComCLIP}} = 
    \lambda_{\text{clip}} \mathcal{L}_{\text{clip}} 
    + \lambda_{\text{mse}} \mathcal{L}_{\text{mse}} 
    + \lambda_{\text{rkd}} \mathcal{L}_{\text{rkd}}.
    \label{eq:total}
\end{equation}
$\mathcal{L}_{\text{clip}}$ preserves \emph{image-text alignment} 
via a properly-tempered contrastive objective; 
$\mathcal{L}_{\text{mse}}$ preserves the \emph{pretrained feature 
manifold} by anchoring the post-projection embedding to its frozen 
pretrained counterpart; and $\mathcal{L}_{\text{rkd}}$ transfers the 
\emph{inter-sample relational structure} of DINOv2 features via 
relational distillation. All three losses operate in the shared 
contrastive space, ensuring they jointly shape the same feature 
geometry. The design follows one principle: \emph{refine relative 
structure} (which samples and texts are close to which) through 
$\mathcal{L}_{\text{clip}}$ and $\mathcal{L}_{\text{rkd}}$, 
\emph{while anchoring absolute geometry} through 
$\mathcal{L}_{\text{mse}}$.

\subsection{Properly-Tempered Contrastive Loss}
\label{sec:method-clip}

We retain the standard contrastive objective (Eq.~\ref{eq:infonce}) 
as the primary post-training signal, but with a critical 
modification: we set the contrastive temperature $\tau$ to a value smaller than that of the CLIP pre-training initialization ($\tau \approx 0.07$, i.e., $1/\tau \approx 14$). 
Specifically, we fix $\tau = 0.01$ (i.e., $1/\tau = 100$) 
throughout post-training. The objective is:
\begin{equation}
    \mathcal{L}_{\text{clip}} = 
    \tfrac{1}{2}\bigl[
        \mathcal{L}_{\text{NCE}}(\bm{u}, \bm{v}; \tau)
      + \mathcal{L}_{\text{NCE}}(\bm{v}, \bm{u}; \tau)
    \bigr], \quad \tau = 0.01,
\end{equation}
where $\bm{u}_i = f_v(x_i)$ and $\bm{v}_i = f_t^0(t_i)$.

\paragraph{Why a small temperature is necessary.}
Consider the image-to-text term for sample $i$, with similarities 
$s_{ij} = \bm{u}_i^\top \bm{v}_j$ and softmax weights 
$p_{ij} = \exp(s_{ij}/\tau) / \sum_{k}\exp(s_{ik}/\tau)$. Its gradient 
with respect to the logits is
\begin{equation}
    \frac{\partial \mathcal{L}_i}{\partial s_{ij}} 
    = \frac{1}{\tau}\bigl(p_{ij} - \mathbb{1}[j=i]\bigr),
    \qquad
    p_{ij} \le \exp\!\Bigl(-\frac{s_{ii}-s_{ij}}{\tau}\Bigr)
    \quad (j \neq i),
    \label{eq:grad}
\end{equation}
so the gradient on a negative pair is bounded by 
$\tfrac{1}{\tau}\exp(-\Delta_{ij}/\tau)$, where 
$\Delta_{ij} = s_{ii}-s_{ij}$ is its margin to the positive, and the 
gradient on the positive is $\tfrac{1}{\tau}\sum_{j\neq i} p_{ij}$. 
A pretrained CLIP already separates matched from mismatched pairs, 
i.e., $\Delta_{ij} > 0$ for most negatives in a post-training batch. 
The temperature then sets which pairs receive gradient, which is the
hardness-aware property of the contrastive
loss~\citep{wang2021understanding}. With a small
$\tau$, $\Delta_{ij}/\tau$ is large for all but the hardest 
negatives, so the gradient on easy pairs vanishes exponentially and 
the pretrained geometry around them is left intact (no forgetting), 
while the update concentrates on the few hard negatives whose margin 
is comparable to $\tau$ (refinement). With the pretraining 
initialization $\tau \approx 0.07$, typical margins are of the same 
order as $\tau$, so a large fraction of already-correct pairs in 
every batch still receives gradient. The encoder is then pushed to 
re-fit the small post-training corpus rather than to correct its 
residual errors, which is the forgetting observed in prior work. 
This argument concerns which pairs are updated, not convergence. It 
is consistent with $\mathcal{L}_{\text{clip}}$ alone improving 
zero-shot accuracy from $61.82$ to $62.90$ at $\tau = 0.01$ and with 
every $1/\tau \geq 50$ improving over the base model 
(Sec.~\ref{sec:exp-tau}), and with a text-anchored geometric 
diagnostic in Appendix~\ref{sec:text_anchor_kl}. 
It is specific to the InfoNCE form of Eq.~\eqref{eq:infonce}; 
objectives whose temperature plays a different role, such as the 
KL-based self-distillation of CLIP-Refine, are not covered by it.

\paragraph{Practical guidance.}
The argument gives a range rather than an optimal value: set $\tau$
well below the pretraining initialization, so that
$\Delta_{ij}/\tau \gg 1$ for most in-batch negatives of the pretrained
model. Empirically, every $50 \le 1/\tau \le 200$ improves zero-shot
accuracy over the base model (Sec.~\ref{sec:exp-tau}), and
$\tau = 0.01$ does so at every batch size from $128$ to $2{,}048$
(Appendix~\ref{app:disentangle}); we transfer it unchanged to
ViT-L/14, SigLIP, COCO Caption and CC12M. We do not derive an optimal
$\tau$ for a given batch size or dataset.

\subsection{MSE Anchoring Loss}
\label{sec:method-mse}

To preserve the intra-modal visual feature manifold within the CLIP pretrained image-text contrastive space---which pure 
contrastive optimization on a small dataset tends to distort by 
overfitting to the post-training image-text correspondences---we 
introduce a feature-level MSE anchoring loss between the 
post-projection embeddings of the trainable encoder $f_v$ and a 
frozen copy $f_v^0$:
\begin{equation}
    \mathcal{L}_{\text{mse}} = 
    \frac{1}{B} \sum_{i=1}^{B} 
    \bigl\| \bm{u}(x_i) - \bm{u}^0(x_i) \bigr\|_2^2.
\end{equation}
By anchoring $\bm{u}$ to its pretrained counterpart $\bm{u}^0$, 
$\mathcal{L}_{\text{mse}}$ provides a stable reference in the 
contrastive space, preventing the encoder from drifting under the 
joint optimization of $\mathcal{L}_{\text{clip}}$ and 
$\mathcal{L}_{\text{rkd}}$.

\subsection{Relational Distillation from DINOv2}
\label{sec:method-rkd}

To enrich the visual structure of CLIP's features beyond what 
image-text pairs supervise, we distill from 
DINOv2~\citep{oquab2023dinov2}, whose self-supervised features encode 
visual structure complementary to CLIP's contrastive pretraining. 
Directly aligning CLIP's features to DINOv2's would force the 
CLIP feature space into DINOv2's purely visual coordinate 
system---catastrophically degrading image-text alignment. 
Following relational knowledge
distillation~\citep{park2019rkd,tung2019sp}, we instead distill the
\emph{relational structure} between samples within a batch.

Let $\bm{u}(x_i)$ and $\bm{g}(x_i)$ denote the 
$\ell_2$-normalized post-projection embeddings of the trainable 
CLIP encoder and the frozen DINOv2 encoder, respectively. We 
compute batch-wise self-similarity matrices and convert them into 
row-wise distributions via a temperature-scaled softmax:
\begin{align}
    S^{\text{clip}}_{ij} &= \bm{u}(x_i)^\top \bm{u}(x_j),
    &
    S^{\text{dino}}_{ij} &= \bm{g}(x_i)^\top \bm{g}(x_j), \\[2pt]
    P^{\text{clip}}_{ij} &= \frac{\exp(S^{\text{clip}}_{ij}/T)}
                                  {\sum_{k}\exp(S^{\text{clip}}_{ik}/T)},
    &
    P^{\text{dino}}_{ij} &= \frac{\exp(S^{\text{dino}}_{ij}/T)}
                                {\sum_{k}\exp(S^{\text{dino}}_{ik}/T)}.
\end{align}
The relational distillation loss is the row-wise KL divergence,
the distributional relation-matching form used in
ReSSL~\citep{zheng2021ressl}:
\begin{equation}
    \mathcal{L}_{\text{rkd}} = 
    \frac{1}{B} \sum_{i=1}^{B} 
    \mathrm{KL}\bigl( P^{\text{dino}}_{i,:} \,\big\|\, P^{\text{clip}}_{i,:} \bigr),
\end{equation}
with $T = 0.1$ and DINOv2 features computed without gradients. 
We compute $\mathcal{L}_{\text{rkd}}$ on the post-projection 
embedding so that all three losses shape the same space; the 
pre-projection alternative performs comparably 
(Appendix~\ref{sec:exp-rkd-position}). In a multi-seed control, 
adding $\mathcal{L}_{\text{rkd}}$ to 
$\mathcal{L}_{\text{clip}} + \mathcal{L}_{\text{mse}}$ significantly 
improves linear probing on ViT-L/14 (Sec.~\ref{sec:exp-rkd-seeds}).

\paragraph{Relation to KUEA.}
KUEA~\citep{uea2024} also distills DINOv2's relational structure
under an $\ell_2$ anchor, but matches kernel values element-wise,
$[k_1(\bm{u}_i,\bm{u}_j)-k_2(\bm{g}_i,\bm{g}_j)]^2$, with a trainable
CLIP-side kernel to calibrate absolute similarity magnitudes.
$\mathcal{L}_{\text{rkd}}$ is parameter-free and, as
$\mathrm{softmax}(\bm{z}{+}c)=\mathrm{softmax}(\bm{z})$, constrains each
sample's relative neighborhood rather than absolute similarities. These
are left to $\mathcal{L}_{\text{mse}}$, so the two terms do not
compete for them, unlike value matching, which pulls them toward
DINOv2's scale. This is a design rationale; Table~\ref{tab::loss_ablation}
compares full recipes, not a single-factor swap.

\subsection{Complementary Effects of the Three Losses}
\label{sec:method-synergy}

$\mathcal{L}_{\text{clip}}$ and $\mathcal{L}_{\text{rkd}}$ are both
row-wise softmax objectives over in-batch similarities, so both act on
relative structure: with a small $\tau$, $\mathcal{L}_{\text{clip}}$
refines how texts are ranked around each image
(Sec.~\ref{sec:method-clip}), and $\mathcal{L}_{\text{rkd}}$ adds
DINOv2's ranking of images around it, a self-normalizing target of the
same kind. A kernel-value term such as KUEA's instead fixes absolute
magnitudes to DINOv2's scale, against the relative structure that
$\mathcal{L}_{\text{clip}}$ sharpens. $\mathcal{L}_{\text{mse}}$, the
only term on absolute position, keeps both from over-shaping the
features. Sec.~\ref{sec:exp-ablation} and
Sec.~\ref{sec:exp-rkd-seeds} test this empirically.

\section{Experiments}
\label{sec:exp-setup}
\subsection{Settings}
\textbf{Pre-trained models.}
We experiment with two student backbones: OpenAI CLIP ViT-B/16
and ViT-L/14. DINOv2 ViT-L/14 (with
registers)~\citep{oquab2023dinov2} serves as the visual teacher
for $\mathcal{L}_{\text{rkd}}$. During post-training, both the
student's text encoder and DINOv2 are kept frozen; only the
student's vision encoder is updated.

\textbf{Post-training datasets.}
We use three datasets spanning different scales and curation
styles: \textbf{CC3M}~\citep{cc3m} and
\textbf{CC12M}~\citep{cc12m} (web-scraped, $3$M and $12$M pairs)
and \textbf{COCO Caption}~\citep{lin2014microsoft} ($591$K
human-annotated pairs).

\textbf{Baselines.}
We compare against (i) the original \textbf{OpenAI CLIP} without
post-training, (ii) \textbf{CLIP-Refine}~\citep{cliprefine2024}
trained on COCO Caption and on CC3M for $1$ epoch, and (iii)
\textbf{KUEA}~\citep{uea2024} trained on ImageNet-1K~\citep{imagenet}
and on CC3M for $2$ epochs (its default schedule), all reproduced
from the official codebases with their original hyperparameters.
Generative-model-based methods (DIVA, un2CLIP, GenHancer) are
compared using their published numbers in
Appendix~\ref{app:generative}.

\textbf{Training details.}
ComCLIP is implemented on top of the official KUEA codebase. We
use AdamW~\citep{adamw} with learning rate $1\!\times\!10^{-6}$,
weight decay $0.1$, and a $410$-step linear warm-up. Each model
is post-trained for $1$ epoch on $8$ A800 GPUs with global batch size
$1{,}024$ and the standard per-sample CLIP data augmentation
inherited from the KUEA codebase. Because augmentation is applied
per sample, the features of the frozen CLIP copy and of DINOv2 are
computed on the fly (forward pass only) rather than cached. The
contrastive temperature is fixed at $\tau = 0.01$, and all three loss
weights are set to $\lambda_{\text{clip}} =
\lambda_{\text{mse}} = \lambda_{\text{rkd}} = 1.0$. The delivered
model is a single CLIP vision encoder with the same architecture and
parameter count as the input model, so inference cost is unchanged.

\textbf{Evaluation.}
We evaluate on five task families:
(i) zero-shot classification averaged over $12$ datasets;
(ii) image-text retrieval (R@1) on MS COCO~\citep{mscoco} and
Flickr30K~\citep{flickr};
(iii) linear probing averaged over $5$ datasets;
(iv) MMVP~\citep{tong2024mmvp} for fine-grained visual
perception; and
(v) downstream VLM evaluation on $8$ benchmarks via
LLaVA-1.5-7B integration. The full list of zero-shot and
linear-probing datasets, along with the VLM benchmarks, is
provided in Appendix~\ref{app:datasets}. MMVP contains only
$N=135$ image pairs, so one pair corresponds to $0.74$ points and
seed-to-seed variation is several points; we therefore base MMVP
claims only on multi-seed comparisons (Sec.~\ref{sec:exp-seeds}).

\begin{table}[t!]
\caption{Comparison of post-training methods on two backbones (ViT-B/16 and ViT-L/14). The ``+'' denotes the application of a post-training method and dataset to the pre-trained OpenAI CLIP. Each method is reported under its strongest validated setting (Sec.~\ref{sec:exp-main}): ComCLIP and CLIP-Refine are trained for $1$ epoch and KUEA for $2$ epochs (its default schedule). Cells with a subscript are means $\pm$ standard deviation over seeds ($3$ seeds; see Table~\ref{tab::seeds}); all other cells are single runs. MMVP has $N=135$ image pairs and a seed standard deviation of up to $\sim$$3$ points, so single-run MMVP differences should not be over-interpreted. The best and second-best results within each backbone are in \textbf{bold} and \underline{underline}; highlighting does not imply statistical significance.}
\label{tab::main_results}
\centering
\resizebox{\textwidth}{!}{
\begin{tabular}{l|c|cc|cc|c|c}
\toprule
\multirow{2}{*}{Method} & Zero-shot & \multicolumn{2}{c|}{MS COCO (R@1)} & \multicolumn{2}{c|}{Flickr30K (R@1)} & Linear Probe & MMVP \\
& CLS Avg & I$\rightarrow$T & T$\rightarrow$I & I$\rightarrow$T & T$\rightarrow$I & Avg & Avg \\
\midrule
\multicolumn{8}{c}{\textbf{Backbone: ViT-B/16}} \\
\midrule
OpenAI CLIP (no post-training)      & 61.82 & 48.16 & 31.47 & 74.70 & 57.10 & 44.70 & 12.59 \\
+ KUEA (ImageNet-1K)                & 62.07 & 48.48 & 31.99 & 75.00 & 57.62 & 44.65 & 11.85 \\
+ KUEA (CC3M)                       & 61.99 & 48.00 & 31.47 & 75.00 & 57.00 & 44.72 & 14.07 \\
+ CLIP-Refine (COCO Caption)       & 62.87\sd{0.11} & \textbf{54.17}\sd{0.28} & \textbf{37.72}\sd{0.07} & \textbf{81.30}\sd{0.40} & \textbf{64.45}\sd{0.21} & 42.28\sd{0.62} & \underline{16.05}\sd{1.14} \\
+ CLIP-Refine (CC3M)                & 61.59 & 41.12 & 33.29 & 66.20 & 60.18 & 44.89 & \textbf{19.26} \\
+ ComCLIP (COCO Caption)           & 62.97 & 52.24 & 35.36 & \underline{79.50} & 61.36 & \underline{47.23} & 15.56 \\
+ ComCLIP (CC3M)                    & \underline{63.00}\sd{0.13} & \underline{52.49}\sd{0.28} & \underline{36.17}\sd{0.02} & 77.80\sd{0.00} & \underline{63.11}\sd{0.09} & \textbf{48.99}\sd{0.35} & 14.81\sd{2.97} \\
+ ComCLIP (CC12M)                   & \textbf{63.04} & 51.58 & 35.56 & 77.10 & 62.42 & 46.56 & 15.56 \\
\midrule
\multicolumn{8}{c}{\textbf{Backbone: ViT-L/14}} \\
\midrule
OpenAI CLIP (no post-training)      & 66.12 & 50.04 & 34.29 & 77.90 & 59.80 & 54.90 & 17.78 \\
+ KUEA (ImageNet-1K)                & 66.89 & 50.88 & 35.66 & 79.50 & 61.18 & \textbf{59.92} & 17.78 \\
+ KUEA (CC3M)                       & 66.36 & 50.30 & 34.68 & 78.00 & 60.02 & 55.82 & 18.52 \\
+ CLIP-Refine (COCO Caption)       & 66.31 & 53.28 & 38.21 & 80.10 & 64.56 & 50.83 & 17.04 \\
+ CLIP-Refine (CC3M)                & -- & -- & -- & -- & -- & -- & 19.01\sd{1.86} \\
+ ComCLIP (COCO Caption)           & 67.09 & \underline{54.32} & \underline{38.79} & \underline{81.80} & 64.98 & \underline{58.28} & 20.74 \\
+ ComCLIP (CC3M)                    & \underline{67.75} & \textbf{54.56} & \textbf{39.40} & \textbf{82.60} & \textbf{65.82} & 56.99 & \textbf{24.20}\sd{0.86} \\
+ ComCLIP (CC12M)                   & \textbf{67.78} & 53.74 & 38.63 & 80.10 & \underline{65.00} & 57.42 & \underline{22.22} \\
\bottomrule
\end{tabular}
}
\vspace{-1em}
\end{table}

\subsection{Main Results}
\label{sec:exp-main}

Table~\ref{tab::main_results} compares ComCLIP with the original
OpenAI CLIP and two post-training baselines,
KUEA~\citep{uea2024} and CLIP-Refine~\citep{cliprefine2024}, on
ViT-B/16 and ViT-L/14. Since different post-training methods favor
different datasets and training budgets, the table reports each
method under its \emph{strongest validated setting} rather than
forcing a single shared configuration. Concretely, CLIP-Refine is
trained on COCO Caption, where it performs best, and additionally on
CC3M; KUEA is trained on ImageNet-1K and on CC3M for $2$ epochs;
ComCLIP and CLIP-Refine use $1$ epoch, since $2$ epochs gave no
consistent improvement (Appendix~\ref{app:epochs}). As a result, the
best-setting rows compare ComCLIP (CC3M) with CLIP-Refine (COCO
Caption), i.e., methods trained on different data; the
dataset-matched rows (both on CC3M for ViT-B/16, both on COCO
Caption for ViT-L/14) are included in the same table and discussed
below.

\textbf{ComCLIP and CLIP-Refine occupy different points of a
trade-off.}
On ViT-B/16, over three seeds, ComCLIP and CLIP-Refine are at parity
on zero-shot classification ($63.00_{\pm0.13}$ vs.\
$62.87_{\pm0.11}$, $p=0.26$), CLIP-Refine is ahead on all four
retrieval metrics, and ComCLIP is far ahead on linear probing
($48.99_{\pm0.35}$ vs.\ $42.28_{\pm0.62}$, $p<0.001$;
Table~\ref{tab::seeds}). CLIP-Refine raises retrieval while reducing
linear-probe accuracy below the original CLIP ($44.70$), whereas
ComCLIP improves it. On ViT-L/14 the same pattern holds for linear
probing (CLIP-Refine $50.83$ vs.\ original $54.90$; ComCLIP
$56.99$--$58.28$), and ComCLIP is also ahead on retrieval.

\textbf{The linear-probe advantage is not an artifact of the training
data.}
Under dataset-matched comparisons, ComCLIP retains a large
linear-probe advantage over CLIP-Refine: with both methods trained on
CC3M on ViT-B/16, $48.99$ vs.\ $44.89$ ($+4.10$), and with both
trained on COCO Caption on ViT-L/14, $58.28$ vs.\ $50.83$ ($+7.45$).

\textbf{KUEA} yields small zero-shot changes (at most $+0.77$); its
highest linear-probe accuracy (ViT-L/14, ImageNet-1K, $59.92$) is
plausibly helped by ImageNet-1K being in the probe suite, and on CC3M
its gain is small ($55.82$).

\textbf{Effect of the post-training corpus.}
Gains are not monotone in corpus size: on ViT-B/16, CC12M matches
CC3M on zero-shot ($63.04$ vs.\ $63.00_{\pm0.13}$) but is lower on
linear probing ($46.56$ vs.\ $48.99_{\pm0.35}$), and ViT-L/14
zero-shot is flat ($67.75$ vs.\ $67.78$). We use CC3M by default
(see Sec.~\ref{sec:conclusion}).

\begin{table}[t!]
\caption{Multi-seed comparison between ComCLIP and CLIP-Refine (mean $\pm$ standard deviation). \textbf{Top:} ViT-B/16, $3$ seeds per method, each method in its best setting (ComCLIP on CC3M, CLIP-Refine on COCO Caption). \textbf{Bottom:} ViT-L/14 MMVP, $3$ seeds per method, both trained on CC3M. $p$: two-sided Welch $t$-test. MMVP has $N=135$ image pairs; on ViT-B/16 the ComCLIP seeds correspond to $\{16, 20, 24\}$ correct pairs.}
\label{tab::seeds}
\centering
\resizebox{\textwidth}{!}{
\begin{tabular}{l|c|cc|cc|c|c}
\toprule
\multirow{2}{*}{Method} & Zero-shot & \multicolumn{2}{c|}{MS COCO (R@1)} & \multicolumn{2}{c|}{Flickr30K (R@1)} & Linear Probe & MMVP \\
& CLS Avg & I$\rightarrow$T & T$\rightarrow$I & I$\rightarrow$T & T$\rightarrow$I & Avg & Avg \\
\midrule
\multicolumn{8}{c}{\textbf{ViT-B/16} ($3$ seeds)} \\
\midrule
CLIP-Refine (COCO Caption) & 62.87\sd{0.11} & 54.17\sd{0.28} & 37.72\sd{0.07} & 81.30\sd{0.40} & 64.45\sd{0.21} & 42.28\sd{0.62} & 16.05\sd{1.14} \\
ComCLIP (CC3M)             & 63.00\sd{0.13} & 52.49\sd{0.28} & 36.17\sd{0.02} & 77.80\sd{0.00} & 63.11\sd{0.09} & 48.99\sd{0.35} & 14.81\sd{2.97} \\
Welch $p$                  & $0.26$ & $2\!\times\!10^{-3}$ & $3\!\times\!10^{-4}$ & $4\!\times\!10^{-3}$ & $3\!\times\!10^{-3}$ & $4\!\times\!10^{-4}$ & $0.56$ \\
\midrule
\multicolumn{8}{c}{\textbf{ViT-L/14} ($3$ seeds, both on CC3M)} \\
\midrule
CLIP-Refine (CC3M) & \multicolumn{6}{c|}{--} & 19.01\sd{1.86} \\
ComCLIP (CC3M)     & \multicolumn{6}{c|}{--} & 24.20\sd{0.86} \\
Welch $p$          & \multicolumn{6}{c|}{--} & $0.025$ \\
\bottomrule
\end{tabular}
}
\vspace{-1em}
\end{table}

\subsection{Multi-seed Statistics}
\label{sec:exp-seeds}

Table~\ref{tab::seeds} reports seed statistics for the comparisons
the paper relies on. On ViT-B/16 (three seeds), the only significant
difference between ComCLIP and CLIP-Refine in favor of ComCLIP is
linear probing (Welch $t=16.3$, $p<0.001$); CLIP-Refine is
significantly ahead on all four retrieval metrics ($p<0.005$); the
zero-shot difference is not significant ($p=0.26$), and on MMVP the two are
indistinguishable ($14.81_{\pm2.97}$ vs.\ $16.05_{\pm1.14}$,
$p=0.56$). On ViT-B/16, the ComCLIP MMVP seeds span
$[11.85, 17.78]$, and every MMVP value in our ViT-B/16 ablations
(Sec.~\ref{sec:exp-ablation-section}) lies within this range; we
therefore draw no conclusions from MMVP on ViT-B/16. On ViT-L/14,
with three seeds per method and both methods trained on CC3M, ComCLIP
improves MMVP over CLIP-Refine ($24.20_{\pm0.86}$ vs.\
$19.01_{\pm1.86}$, Welch $t=4.4$, $p=0.025$); this is
the only MMVP claim we make. We note that generative enhancement
methods report higher MMVP on the same backbone
(Appendix~\ref{app:generative}).

\begin{table}[t!]
\caption{LLaVA-1.5-7B with the vision encoder swapped for ComCLIP ViT-L/14 (post-trained on CC3M), KUEA ViT-L/14 (ImageNet-1K), CLIP-Refine ViT-L/14 (COCO Caption), or the original OpenAI CLIP ViT-L/14; the LLM and projector are kept frozen. `*' indicates official KUEA weights~\citep{uea2024}. Because the benchmarks differ in scale by more than $60\times$ (e.g., MME-p vs.\ RefCOCOg), we do not report a raw arithmetic mean. Instead, \textbf{Rel.\,$\Delta$} is the mean per-benchmark relative change w.r.t.\ OpenAI CLIP (\%), \textbf{Ceil.\,$\Delta$} is the mean change after dividing each benchmark by its maximum attainable score (points), and \textbf{W/T/L} counts benchmarks improved/tied/degraded w.r.t.\ OpenAI CLIP. The best result in each benchmark column is in \textbf{bold}.}
\label{tab::llava_benchmark}
\centering
\resizebox{\textwidth}{!}{
\begin{tabular}{c|ccc|ccccccccc}
\toprule
Method & Rel.\,$\Delta$ & Ceil.\,$\Delta$ & W/T/L & AI2D & POPE & RefCOCOg & V$^*$ & MME-c & MME-p & SQA-Img & OCRBench & TextVQA \\
\midrule
OpenAI CLIP & -- & -- & -- & \textbf{52.75} & 85.20 & \textbf{20.67} & \textbf{41.36} & \textbf{320.00} & 1350.79 & 67.63 & 27.80 & 34.57 \\
KUEA\textsuperscript{*} & $-1.23$ & $-0.51$ & 3/0/6 & 52.30 & 85.46 & 20.50 & 38.74 & 317.14 & 1350.82 & 66.68 & 27.40 & 34.69 \\
CLIP-Refine & $-0.95$ & $-0.37$ & 3/0/6 & 52.20 & \textbf{85.57} & 20.61 & 39.79 & 308.21 & 1345.97 & 67.72 & 27.60 & 34.84 \\
ComCLIP & $+0.02$ & $+0.04$ & 4/2/3 & \textbf{52.75} & 85.14 & 20.63 & \textbf{41.36} & 307.50 & \textbf{1373.47} & \textbf{67.82} & \textbf{28.20} & \textbf{34.89} \\
\bottomrule
\end{tabular}
}
\end{table}

\subsection{Downstream VLM Compatibility}
\label{sec:exp-llava}

To verify that the refined encoder remains a drop-in replacement,
we replace LLaVA-1.5-7B's vision encoder (originally CLIP
ViT-L/14-336) with ComCLIP, KUEA, CLIP-Refine and the
original OpenAI CLIP (all ViT-L/14-224), while keeping the LLM
and cross-modal projector strictly frozen
(Table~\ref{tab::llava_benchmark}). Because the benchmarks differ in
scale by more than an order of magnitude, we aggregate with
scale-normalized measures rather than a raw arithmetic mean. Under
these measures, ComCLIP produces \emph{no net change} relative to the
original CLIP: the mean per-benchmark relative change is
$+0.02\%$, the ceiling-normalized change is $+0.04$ points, and it
improves $4$, ties $2$, and degrades $3$ of the $9$ scores;
excluding the two MME scores, the mean change is $+0.12$ points
($+0.34\%$). The largest individual movements are on MME, with
perception up by $22.68$ and cognition down by $12.50$, i.e., by more
than twice as much in relative terms ($+1.7\%$ vs.\ $-3.9\%$). By
contrast, KUEA and CLIP-Refine reduce the aggregate ($-1.23\%$ and
$-0.95\%$ relative, each degrading $6$ of $9$ scores). We read these
results as evidence of compatibility: ComCLIP's refinement of the
vision encoder does not break the frozen projector and LLM trained on
the original CLIP, but it does not by itself improve downstream VLM
accuracy without re-alignment.

\subsection{Ablation Study}
\label{sec:exp-ablation-section}

Ablations are run on ViT-B/16 with CC3M. Unless a seed standard
deviation is given, entries are single runs; for reference, the
ViT-B/16 seed standard deviations of ComCLIP are $0.13$ (zero-shot),
$0.35$ (linear probe) and $2.97$ (MMVP). Every ViT-B/16 MMVP value in
this section lies within the seed range of a single configuration, so
we report MMVP for completeness only.

\subsubsection{Loss Decoupling}
\label{sec:exp-ablation}
\begin{table}[t!]
\caption{Loss decoupling ablation on ViT-B/16, post-trained on CC3M with batch size $128 \times 8$. All configurations of $\{\mathcal{L}_\text{clip}, \mathcal{L}_\text{mse}, \mathcal{L}_\text{rkd}\}$ are trained for $1$ epoch. KUEA and KUEA\,+\,$\mathcal{L}_\text{clip}$ use KUEA's default $2$-epoch schedule and hyperparameters, which were not re-tuned (loss weight, learning rate, anchor weight) after adding $\mathcal{L}_\text{clip}$. Temperature is fixed at $\tau = 0.01$ throughout. Entries are single runs except where a seed standard deviation is given. Best results per column (except MMVP) in \textbf{bold}. All MMVP values ($N=135$) lie within the $3$-seed range of ComCLIP ($[11.85, 17.78]$), so MMVP is not highlighted and no conclusion is drawn from it.}
\label{tab::loss_ablation}
\centering
\resizebox{\textwidth}{!}{
\begin{tabular}{l|c|cc|cc|c|c}
\toprule
\multirow{2}{*}{Method} & Zero-shot & \multicolumn{2}{c|}{MS COCO (R@1)} & \multicolumn{2}{c|}{Flickr30K (R@1)} & Linear Probe & MMVP \\
& 12 datasets Avg & I$\rightarrow$T & T$\rightarrow$I & I$\rightarrow$T & T$\rightarrow$I & 5 datasets Avg & Avg \\
\midrule
OpenAI CLIP (no post-training) & 61.82 & 48.16 & 31.47 & 74.70 & 57.10 & 44.70 & 12.59 \\
\midrule
+ KUEA  & 61.99 & 48.00 & 31.47 & 75.00 & 57.00 & 44.72 & 14.07 \\
+ (KUEA + $\mathcal{L}_\text{clip}$)  & 62.24 & 48.84 & 34.20 & 75.30 & 60.80 & 45.25 & 13.33 \\
\midrule
+ $\mathcal{L}_\text{clip}$  & 62.90 & 51.92 & \textbf{38.28} & 76.80 & \textbf{64.56} & 44.04 & 14.07 \\
+ $\mathcal{L}_\text{mse}$   & 61.82 & 48.18 & 31.45 & 74.70 & 57.12 & 44.79 & 11.85 \\
+ $\mathcal{L}_\text{rkd}$   & 56.64 & 43.30 & 25.57 & 69.30 & 49.66 & 44.42 & 17.78 \\
\midrule
+ $(\mathcal{L}_\text{mse} + \mathcal{L}_\text{rkd})$   & 61.36 & 49.98 & 31.53 & 75.40 & 58.08 & 45.50 & 15.56 \\
+ $(\mathcal{L}_\text{clip} + \mathcal{L}_\text{mse})$  & \textbf{63.35} & 51.64 & 37.03 & 77.00 & 64.34 & 48.60 & 14.07 \\
+ $(\mathcal{L}_\text{clip} + \mathcal{L}_\text{rkd})$  & 62.49 & 51.50 & 36.67 & 77.30 & 63.96 & 47.15 & 14.07 \\
\midrule
+ ComCLIP ($\mathcal{L}_\text{clip} + \mathcal{L}_\text{mse} + \mathcal{L}_\text{rkd}$) & 63.15 & \textbf{52.20} & 36.18 & \textbf{77.80} & 63.22 & \textbf{48.74} & 14.81\sd{2.97} \\
\bottomrule
\end{tabular}
}
\vspace{-1em}
\end{table}

Table~\ref{tab::loss_ablation} reports all single-loss, pairwise, and
full configurations of $\{\mathcal{L}_{\text{clip}},
\mathcal{L}_{\text{mse}}, \mathcal{L}_{\text{rkd}}\}$ (Finding~2).
\textbf{Each loss moves a different group of metrics.}
$\mathcal{L}_{\text{clip}}$ alone improves zero-shot classification
($+1.08$) and retrieval but leaves linear probing essentially
unchanged ($44.04$ vs.\ $44.70$). $\mathcal{L}_{\text{mse}}$ alone is
a no-op (differences within $0.1$). $\mathcal{L}_{\text{rkd}}$ alone
severely degrades zero-shot classification ($-5.18$) and retrieval
($-7.44$ on Flickr30K T$\rightarrow$I): distilling DINOv2's relational
structure without an image-text signal pulls the embedding away from
the text space.
\textbf{Combinations.}
$\mathcal{L}_{\text{clip}} \!+\! \mathcal{L}_{\text{mse}}$ achieves
the highest zero-shot accuracy ($63.35$) and a large linear-probe gain
over $\mathcal{L}_{\text{clip}}$ alone ($48.60$ vs.\ $44.04$):
anchoring the absolute geometry lets the contrastive signal improve
alignment without eroding transferable features.
$\mathcal{L}_{\text{clip}} \!+\! \mathcal{L}_{\text{rkd}}$ improves
linear probing less ($47.15$), and
$\mathcal{L}_{\text{mse}} \!+\! \mathcal{L}_{\text{rkd}}$ leaves
zero-shot accuracy below the baseline. The full objective reaches the
highest linear-probe accuracy ($48.74$) with zero-shot accuracy
($63.15$) within $\sim$$1.5$ seed standard deviations of the best
variant. Adding properly-tempered $\mathcal{L}_{\text{clip}}$ to KUEA,
without re-tuning KUEA's hyperparameters, also improves its zero-shot
accuracy ($+0.25$) and retrieval (up to $+3.80$ on Flickr30K
T$\rightarrow$I), so the temperature finding is not tied to ComCLIP's
other components; ComCLIP remains ahead of
KUEA\,+\,$\mathcal{L}_{\text{clip}}$ on zero-shot ($+0.91$) and linear
probing ($+3.49$), a comparison between recipes rather than a
single-factor ablation (see the caption of
Table~\ref{tab::loss_ablation}).

\subsubsection{Contribution of \texorpdfstring{$\mathcal{L}_{\text{rkd}}$}{L\_rkd}}
\label{sec:exp-rkd-seeds}

Because the single-run effect of $\mathcal{L}_{\text{rkd}}$ above is
small, we isolate it with a $3$-seed control that adds
$\mathcal{L}_{\text{rkd}}$ to $\mathcal{L}_{\text{clip}} +
\mathcal{L}_{\text{mse}}$ (Appendix~\ref{app:rkd},
Table~\ref{tab::rkd_seeds}). On ViT-L/14, $\mathcal{L}_{\text{rkd}}$
significantly improves linear probing ($56.36_{\pm0.23} \rightarrow
57.12_{\pm0.14}$, Welch $t=4.9$, $p \approx 0.013$); on
ViT-B/16 the effect has the same sign but is not significant
($48.54_{\pm0.10} \rightarrow 48.94_{\pm0.41}$, $p \approx 0.23$),
consistent with a benefit that grows with student capacity. It is not
an all-metric gain: on ViT-L/14, zero-shot accuracy is unchanged
($67.76 \rightarrow 67.75$) and retrieval changes are small and mixed
(e.g., COCO T$\rightarrow$I $40.37 \rightarrow 39.40$). We therefore
retain $\mathcal{L}_{\text{rkd}}$ on the basis of linear-probe
transferability only. The benefit also holds with a DINOv2-Base
teacher, but our comparison does not single out a
teacher size (Appendix~\ref{app:rkd}).

\subsubsection{Effect of Contrastive Temperature}
\label{sec:exp-tau}
\begin{table}[t!]
\caption{Ablation study on the initialization of the temperature parameter ($\tau$) in the CLIP loss. We conduct this experiment \textbf{exclusively using the CLIP loss} on the pre-trained OpenAI CLIP ViT-B/16 model to isolate the impact of $\tau$. We conduct post-training on the CC3M dataset. The global batch size is fixed at $128 \times 8$ (1024) across all settings. The best results are highlighted in \textbf{bold}. MMVP ($N=135$, single runs) is reported for completeness only: its variation across rows is within seed-to-seed variance (Table~\ref{tab::seeds}), so it is not highlighted and not used to select $\tau$.}
\label{tab::logit_scale_ablation}
\centering
\resizebox{\textwidth}{!}{
\begin{tabular}{c|c|cc|cc|c|c}
\toprule
\multirow{2}{*}{$\tau$} & Zero-shot & \multicolumn{2}{c|}{MS COCO (R@1)} & \multicolumn{2}{c|}{Flickr30K (R@1)} & Linear Probe & MMVP \\
& 12 datasets Avg & I$\rightarrow$T & T$\rightarrow$I & I$\rightarrow$T & T$\rightarrow$I & 5 datasets Avg & Avg \\
\midrule
OpenAI CLIP (no post-training) & 61.82 & 48.16 & 31.47 & 74.70 & 57.10 & 44.70 & 12.59 \\
\midrule
0.07  & 57.90 & 39.10 & 34.12 & 65.70 & 58.86 & 43.95 & 16.30 \\
0.03  & 61.51 & 48.94 & 37.01 & 74.50 & 63.54 & 43.12 & 13.33 \\
0.02  & 62.57 & 51.54 & \textbf{38.33} & 76.70 & \textbf{65.00} & 43.98 & 13.33 \\
0.01 & \textbf{62.90} & \textbf{52.50} & 38.26 & \textbf{76.80} & 64.56 & 44.04 & 14.07 \\
0.007 & 62.72 & 50.72 & 37.74 & 75.90 & 64.28 & 44.53 & 17.78 \\
0.005 & 62.36 & 49.38 & 37.48 & 75.10 & 64.10 & \textbf{45.90} & 14.81 \\
\bottomrule
\end{tabular}
}
\vspace{-1em}
\end{table}

To examine Finding~1, we train with $\mathcal{L}_{\text{clip}}$ only
and sweep $\tau$ (Table~\ref{tab::logit_scale_ablation}).
\textbf{The pretraining temperature reproduces catastrophic
forgetting.} At $\tau = 0.07$ (the CLIP pretraining initialization),
zero-shot accuracy drops by $3.92$ and Flickr30K I$\rightarrow$T by
$9.00$, consistent with the gradient analysis in
Sec.~\ref{sec:method-clip}.
\textbf{Sufficiently small $\tau$ enables successful post-training.}
For every tested $50 \le 1/\tau \le 200$, contrastive post-training
\emph{improves} zero-shot accuracy (all in $[62.36, 62.90]$) and
retrieval: the \emph{magnitude} of $\tau$, not its exact value,
matters. We adopt $\tau = 0.01$ (best zero-shot, best or near-best
retrieval; MMVP not used).
\textbf{Batch size and learnability.}
With $\tau = 0.01$, larger batches still help ($+1.03$ zero-shot from
$128$ to $2{,}048$), but far less than the $-3.92$ caused by
$\tau = 0.07$, so limited negatives alone do not explain the failures
reported by CLIP-Refine~\citep{cliprefine2024}; fixed and learnable
$\tau$ differ by at most $0.05$ (Appendix~\ref{app:disentangle}).

\section{Conclusion and Limitations}
\label{sec:conclusion}

We presented \textbf{ComCLIP}, a lightweight single-epoch CLIP
post-training recipe at a fixed text encoder. We showed that, for
InfoNCE, the forgetting attributed to contrastive post-training is
driven primarily by the temperature magnitude rather than by small
batches: with a small $\tau$, already-separated pairs receive
vanishing gradient and contrastive post-training improves the
pretrained model. Relative to CLIP-Refine, ComCLIP is at parity on
zero-shot classification, significantly better on linear-probe
transferability (on both backbones and under dataset-matched data),
better on MMVP on ViT-L/14, and behind on retrieval; inside
LLaVA-1.5-7B it is a drop-in replacement with no net change.

\paragraph{Limitations and Future Work.}
\textit{(i)} ComCLIP targets CLIP-like models with a shared
contrastive image-text space; extending it to vision encoders or VLMs
without such a space is future work. The temperature analysis covers
InfoNCE only; it does
not directly apply to the KL-based self-distillation of CLIP-Refine
or to SigLIP's sigmoid loss (Appendix~\ref{post_siglip} applies our
InfoNCE objective to a SigLIP backbone).
\textit{(ii)} MMVP has only $135$ pairs and large seed variance; our
MMVP gain over CLIP-Refine is significant only on ViT-L/14, and
generative methods report substantially higher MMVP
(Appendix~\ref{app:generative}). ComCLIP is not a method for
maximizing fine-grained perception.
\textit{(iii)} Our downstream evaluation uses a single VLM
configuration (LLaVA-1.5-7B, frozen projector and LLM); other VLM
architectures would better establish generality. Without re-aligning
the projector, ComCLIP preserves
but does not improve downstream VLM accuracy; whether re-alignment
converts its linear-probe gains into VLM gains is untested.
\textit{(iv)} Performance is not monotone in corpus size
(Sec.~\ref{sec:exp-main}); whether this reflects single-epoch
dynamics, data quality, or a capacity limit is open.
\textit{(v)} Training runs a frozen CLIP copy and a DINOv2 teacher
forward at every step, and we report no per-component compute
breakdown; the efficiency we claim is a single epoch on a small
corpus with unchanged inference cost.

\section*{Acknowledgments}
This work is supported by National Natural Science Foundation of China(U22B2017), and International Cooperation Foundation of Hubei Province, China (2024EHA032).

\newpage
{
  \small
  \bibliographystyle{bib}
  \bibliography{reference}
}
\newpage
\appendix
\section{Experiments Settings}
\label{app:datasets}
\paragraph{Datasets Settings}
For zero-shot evaluation, we use the following datasets: ImageNet-1K, CIFAR-10/100~\citep{cifar}, Caltech101~\citep{caltech}, FER2013~\citep{fer}, OxfordPets~\citep{pets}, DTD~\citep{dtd}, RESISC45~\citep{resisc}, 
EuroSAT~\citep{eurosat}, PCAM~\citep{pcam}, ImageNet-Sketch~\citep{insketch}, ImageNet-O~\citep{ino}. We evaluate the model’s linear probing performance on five datasets: ImageNet-1K, SVHN~\cite{netzer2011reading}, GTSRB~\cite{Stallkamp-IJCNN-2011}, CLEVR Distance, and CLEVR Counts~\cite{johnson2017clevr}. For multimodal large language model evaluation, we select the following benchmarks: AI2D~\cite{kembhavi2016diagram}, POPE~\cite{li2023evaluating}, RefCOCOg~\cite{mao2016generation}, VSTAR~\cite{wu2024v}, MME~\cite{fu2023mme}, SQA-Img~\cite{lu2022learn}, OCRBench~\cite{liu2024ocrbench}, and TextVQA~\cite{singh2019towards}.

\section{Post-training Performance Results based on SigLIP}\label{post_siglip}

\begin{table}[h!]
\caption{Performance evaluation of post-training methods based on the SigLIP backbone. We compare our proposed method (\textbf{ComCLIP}) against the pre-trained SigLIP ViT-SO400M/14 baseline and KUEA (post-trained on ImageNet-1K). All methods are evaluated on the CC3M dataset for post-training, except for the KUEA baseline which is evaluated on its original setting. `*' denotes results reported by the original authors\citep{uea2024}. The best results in each column are highlighted in \textbf{bold}.}
\label{tab::siglip_results}
\centering
\resizebox{\textwidth}{!}{
\begin{tabular}{c|c|cc|cc|c|c}
\toprule
\multirow{2}{*}{Method} & Zero-shot & \multicolumn{2}{c|}{MS COCO (R@1)} & \multicolumn{2}{c|}{Flickr30K (R@1)} & Linear Probe & MMVP \\
& CLS Avg & I$\rightarrow$T & T$\rightarrow$I & I$\rightarrow$T & T$\rightarrow$I & Avg & Avg \\
\midrule
SigLIP (no post-training) & 71.24 & 68.16 & 52.20 & 88.90 & 74.80 & 70.87 & 39.26 \\
KUEA (ImageNet-1K)\textsuperscript{*}       & 71.67 & - & - & - & - & - & - \\
\textbf{ComCLIP (CC3M)}   & \textbf{72.15} & \textbf{70.06} & \textbf{53.05} & \textbf{89.50} & \textbf{75.78} & \textbf{70.88} & \textbf{41.48} \\
\bottomrule
\end{tabular}
}
\end{table}

\begin{wrapfigure}{r}{0.5\textwidth}
    \centering
    \vspace{-15pt} 
    \includegraphics[width=0.5\textwidth]{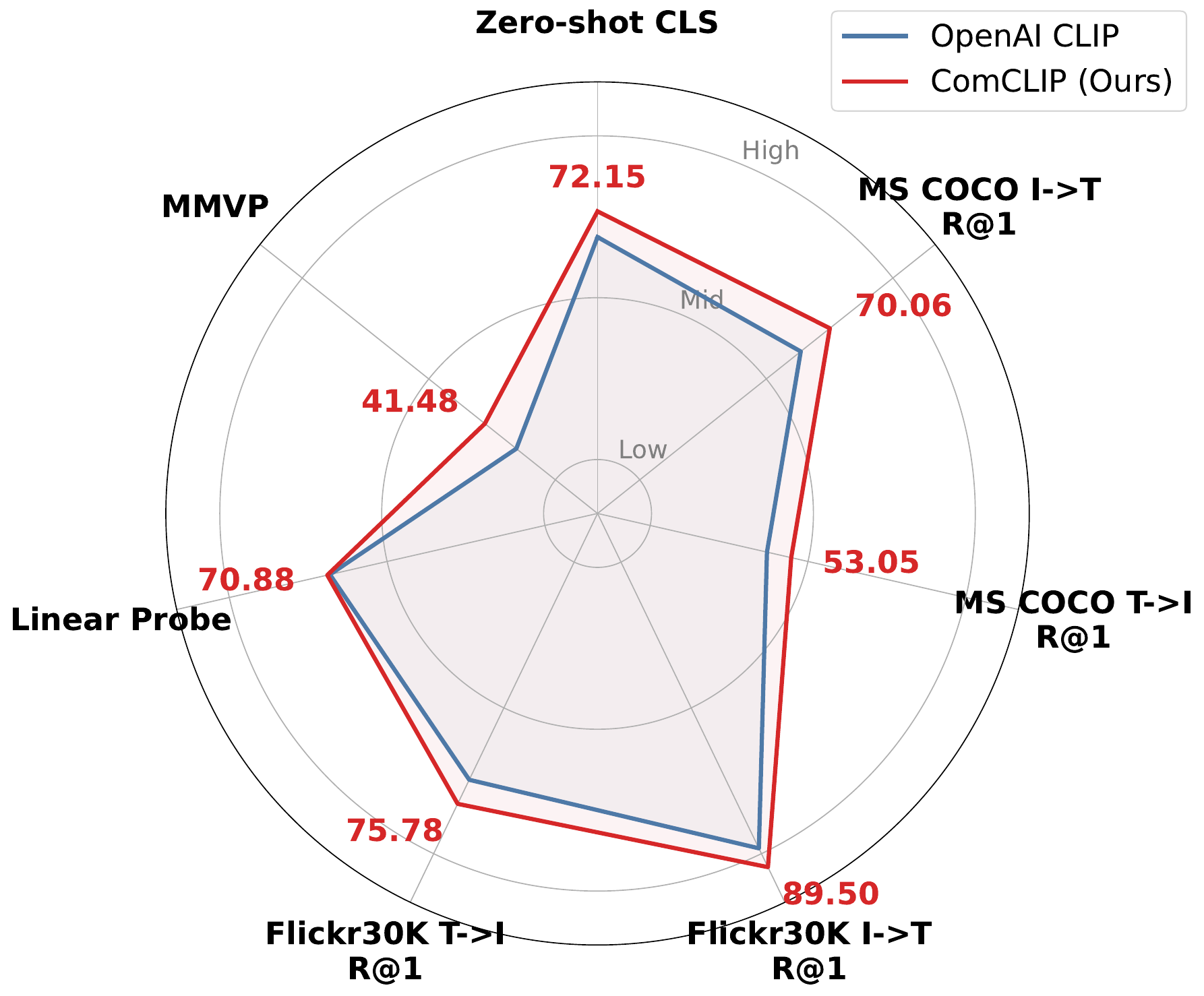}
    \caption{Multi-dimensional performance comparison on SigLIP ViT-SO400M/14.}
    \label{fig:radar_siglip}
    \vspace{-8pt} 
\end{wrapfigure}
\textbf{Applying ComCLIP to a SigLIP backbone.} We apply ComCLIP to the pre-trained SigLIP ViT-SO400M/14, post-training on CC3M for one epoch with $\tau$ fixed at $0.01$ (Table~\ref{tab::siglip_results}, single run). Note that this experiment uses our InfoNCE contrastive objective on a SigLIP backbone; it does not test whether the temperature finding transfers to SigLIP's sigmoid loss, which we leave to future work. Compared with the pre-trained SigLIP and with KUEA (zero-shot result reported in the original paper), ComCLIP improves zero-shot classification ($71.24 \rightarrow 72.15$) and retrieval, and leaves linear probing unchanged ($70.87 \rightarrow 70.88$); the MMVP change ($39.26 \rightarrow 41.48$, three pairs) is within the seed variance we observe elsewhere. Figure~\ref{fig:radar_siglip} visualizes these results.

\section{Disentangling Temperature from Batch Size and Learnability}
\label{app:disentangle}
\begin{table}[t!]
\caption{Ablation study on the effect of batch size during post-training with the CLIP loss. Experiments are conducted on the pre-trained OpenAI CLIP ViT-B/16 model, utilizing the CC3M dataset exclusively for post-training. Only the CLIP loss is applied to update the visual encoder $(\lambda_\text{clip}=1.0, \lambda_\text{mse}=0.0, \lambda_\text{rkd}=0.0)$. The batch sizes are denoted as $(per\text{-}GPU \ batch \ size \times number\ of\ GPUs)$. The best results in each column are highlighted in \textbf{bold}. MMVP ($N=135$, single runs) varies within seed-to-seed variance (Table~\ref{tab::seeds}) and is not highlighted.}
\label{tab::batchsize_ablation}
\centering
\resizebox{\textwidth}{!}{
\begin{tabular}{c|c|cc|cc|c|c}
\toprule
\multirow{2}{*}{Batch Size} & Zero-shot & \multicolumn{2}{c|}{MS COCO (R@1)} & \multicolumn{2}{c|}{Flickr30K (R@1)} & Linear Probe & MMVP \\
& 12 datasets Avg & I$\rightarrow$T & T$\rightarrow$I & I$\rightarrow$T & T$\rightarrow$I & 5 datasets Avg & Avg \\
\midrule
$16 \times 8$  & 62.00 & 50.56 & 37.26 & 76.10 & 63.22 & 42.15 & 15.56 \\
$32 \times 8$  & 62.50 & 51.26 & 37.79 & 76.50 & 64.16 & 42.98 & 11.85 \\
$64 \times 8$  & 62.65 & 51.44 & 38.10 & 76.50 & 64.12 & 43.65 & 14.81 \\
$128 \times 8$ & 62.69 & 52.50 & \textbf{38.26} & 77.40 & 64.52 & 43.68 & 15.56 \\
$256 \times 8$ & \textbf{63.03} & \textbf{52.52} & 38.22 & \textbf{77.80} & \textbf{64.80} & \textbf{44.63} & 18.52 \\
\bottomrule
\end{tabular}
}
\vspace{-1em}
\end{table}

\begin{table}[t!]
\caption{Ablation study on the learnability of $\tau$ and the impact of different post-training datasets (COCO Caption vs. CC3M). All experiments are conducted on the pre-trained OpenAI CLIP ViT-B/16 model, utilizing exclusively the CLIP loss $(\lambda_\text{clip}=1.0, \lambda_\text{mse}=0.0, \lambda_\text{rkd}=0.0)$. During post-training, only the parameters of the visual encoder are updated while the text encoder remains strictly frozen. The best results in each metric are highlighted in \textbf{bold} (MMVP, single runs within seed-to-seed variance, is not highlighted).}
\label{tab::learnability_dataset_ablation}
\centering
\resizebox{\textwidth}{!}{
\begin{tabular}{c|c|c|c|cc|cc|c|c}
\toprule
\multirow{2}{*}{Dataset} & \multirow{2}{*}{Learnable} & \multirow{2}{*}{Batch Size} & Zero-shot & \multicolumn{2}{c|}{MS COCO (R@1)} & \multicolumn{2}{c|}{Flickr30K (R@1)} & Linear Probe & MMVP \\
& & & 12 datasets Avg & I$\rightarrow$T & T$\rightarrow$I & I$\rightarrow$T & T$\rightarrow$I & 5 datasets Avg & Avg \\
\midrule
\multirow{2}{*}{COCO Caption} & Yes & $128 \times 8$ & 62.25 & 49.86 & 34.74 & \textbf{77.20} & 61.38 & \textbf{45.30} & 14.81 \\
                               & No  & $128 \times 8$ & 62.25 & 49.86 & 34.74 & \textbf{77.20} & 61.40 & 45.27 & 14.81 \\
\midrule
\multirow{2}{*}{CC3M}          & Yes & $128 \times 8$ & \textbf{62.95} & 52.22 & 38.22 & 76.90 & 64.48 & 43.42 & 14.07 \\
                               & No  & $128 \times 8$ & 62.90 & \textbf{52.50} & \textbf{38.26} & 76.80 & \textbf{64.56} & 44.04 & 14.07 \\
\bottomrule
\end{tabular}
}
\vspace{-1em}
\end{table}

To rule out two alternative explanations of Finding~1---that the
underlying issue lies with batch size or with the learnability of
$\tau$---we run two controlled studies using only
$\mathcal{L}_{\text{clip}}$ with $\tau = 0.01$ on ViT-B/16.

\textbf{Batch size.}
CLIP-Refine~\citep{cliprefine2024} attributes the failure of
contrastive post-training mainly to insufficient negatives, reporting
that even batch size $2{,}048$ cannot recover CLIP's performance. With
$\tau = 0.01$, contrastive post-training improves zero-shot accuracy
over the original CLIP at every tested batch size from $128$ to
$2{,}048$ (Table~\ref{tab::batchsize_ablation}). Larger batches still
provide gains ($+1.03$ zero-shot from $128$ to $2{,}048$), but these
are much smaller than the degradation caused by an improperly set
$\tau$ ($-3.92$). Temperature calibration is therefore a critical and
previously under-emphasized factor in InfoNCE post-training, and
limited negatives alone do not explain the reported failures.

\textbf{Learnability.}
On COCO Caption and CC3M, with $\tau$ initialized at $0.01$
(Table~\ref{tab::learnability_dataset_ablation}), fixed and learnable
$\tau$ yield near-identical results (within $0.05$ on zero-shot):
initialized at a sufficiently small value, $\tau$ does not drift back
towards the pretraining value. The critical factor is the magnitude of
$\tau$, not whether it remains learnable.

\section{Contribution of \texorpdfstring{$\mathcal{L}_{\text{rkd}}$}{L\_rkd} and Choice of Teacher}
\label{app:rkd}
\begin{table}[t!]
\caption{Contribution of $\mathcal{L}_\text{rkd}$ and choice of teacher. \textbf{Top:} adding $\mathcal{L}_\text{rkd}$ (DINOv2-Large teacher) to $\mathcal{L}_\text{clip}+\mathcal{L}_\text{mse}$, linear probe (5-dataset average), mean $\pm$ standard deviation over $3$ seeds, CC3M, $1$ epoch; $p$: two-sided Welch $t$-test. \textbf{Bottom:} teacher choice for $\mathcal{L}_\text{rkd}$ (single runs, CC3M, $1$ epoch). The ViT-B/16 teacher runs use the same configuration as Table~\ref{tab::loss_ablation}.}
\label{tab::rkd_seeds}
\centering
\small
\begin{tabular}{l|ccc}
\toprule
\multicolumn{4}{c}{\textbf{Linear probe with vs.\ without $\mathcal{L}_\text{rkd}$} ($3$ seeds)} \\
\midrule
Backbone & $\mathcal{L}_\text{clip}+\mathcal{L}_\text{mse}$ & $+\,\mathcal{L}_\text{rkd}$ (ComCLIP) & Welch $p$ \\
\midrule
ViT-B/16 & $48.54_{\pm0.10}$ & $48.94_{\pm0.41}$ & $0.23$ \\
ViT-L/14 & $56.36_{\pm0.23}$ & $57.12_{\pm0.14}$ & $0.013$ \\
\midrule
\multicolumn{4}{c}{\textbf{Teacher for $\mathcal{L}_\text{rkd}$} (single runs)} \\
\midrule
Configuration & \multicolumn{2}{c}{Zero-shot (12 avg)} & Linear Probe (5 avg) \\
\midrule
ViT-B/16, $\mathcal{L}_\text{clip}+\mathcal{L}_\text{mse}$ (no teacher) & \multicolumn{2}{c}{63.35} & 48.60 \\
ViT-B/16, $+\,\mathcal{L}_\text{rkd}$, MAE-Large      & \multicolumn{2}{c}{63.51} & 46.99 \\
ViT-B/16, $+\,\mathcal{L}_\text{rkd}$, DINOv2-Base    & \multicolumn{2}{c}{62.87} & 49.28 \\
ViT-B/16, $+\,\mathcal{L}_\text{rkd}$, DINOv2-Large   & \multicolumn{2}{c}{63.15} & 48.74 \\
\midrule
ViT-L/14, $+\,\mathcal{L}_\text{rkd}$, DINOv2-Base    & \multicolumn{2}{c}{--} & 56.91 \\
ViT-L/14, $+\,\mathcal{L}_\text{rkd}$, DINOv2-Large   & \multicolumn{2}{c}{67.75} & 56.99 \\
\bottomrule
\end{tabular}
\vspace{-1em}
\end{table}

Table~\ref{tab::rkd_seeds} (top) gives the $3$-seed control discussed
in Sec.~\ref{sec:exp-rkd-seeds}. The teacher comparison (bottom) does
not single out a teacher. On ViT-B/16, DINOv2-Base gives the highest
linear-probe accuracy ($49.28$) and MAE-Large the highest zero-shot
accuracy ($63.51$) in single runs; on ViT-L/14, DINOv2-Base and
DINOv2-Large give nearly identical linear probing ($56.91$ vs.\
$56.99$). The benefit of $\mathcal{L}_{\text{rkd}}$ is thus present
even when teacher and student have matched capacity. We use
DINOv2-Large for both backbones for uniformity. These results support
the component but do not by themselves select a teacher size, and a
cheaper DINOv2-Base teacher is a reasonable alternative.

\section{Where to Apply RKD}
\label{sec:exp-rkd-position}
\begin{table}[h!]
\caption{Investigation of the alignment position for DINOv2 rkd loss features on ViT-B/16. We conduct post-training on the CC3M dataset. ComCLIP-bp denotes that the visual features of the student model used for rkd loss calculation are extracted prior to the projection layer. ComCLIP represents our default configuration, where visual features are projected into the joint image-text embedding space via the original projection layer. Single runs except where a seed standard deviation is given; the MMVP difference is within seed-to-seed variance (Table~\ref{tab::seeds}) and is not highlighted.}
\label{tab::rkd_ablation}
\centering
\resizebox{\textwidth}{!}{
\begin{tabular}{c|c|cc|cc|c|c}
\toprule
\multirow{2}{*}{Method} & Zero-shot & \multicolumn{2}{c|}{MS COCO (R@1)} & \multicolumn{2}{c|}{Flickr30K (R@1)} & Linear Probe & MMVP \\
& 12 datasets Avg & I$\rightarrow$T & T$\rightarrow$I & I$\rightarrow$T & T$\rightarrow$I & 5 datasets Avg & Avg \\
\midrule

OpenAI CLIP (no post-training) & 61.82 & 48.16 & 31.47 & 74.70 & 57.10 & 44.70 & 12.59 \\
\midrule
ComCLIP-bp & \textbf{63.19} & 51.44 & \textbf{36.54} & 77.00 & \textbf{63.50} & 48.23 & 13.33 \\
\midrule
ComCLIP & 63.15 & \textbf{52.20} & 36.18 & \textbf{77.80} & 63.22 & \textbf{48.74} & 14.81\sd{2.97} \\

\bottomrule
\end{tabular}
}
\end{table}

We additionally investigate where DINOv2 distillation should be 
applied along the visual encoder. ComCLIP's default configuration 
applies $\mathcal{L}_{\text{rkd}}$ on the post-projection 
embedding $\bm{u}(x_i)$ (the same space used by 
$\mathcal{L}_{\text{clip}}$). A natural alternative, denoted 
\textbf{ComCLIP-bp}, applies it on the pre-projection 
$[\text{CLS}]$ feature $\bm{h}(x_i)$, which retains the full 
pretrained visual semantics before dimensionality reduction.

Table~\ref{tab::rkd_ablation} compares the two choices on 
ViT-B/16 (CC3M, 1 epoch). The two variants perform comparably: 
differences on zero-shot classification, retrieval and linear 
probing are within $\sim$$0.8$ points, and the MMVP difference is 
within seed-to-seed variance (Table~\ref{tab::seeds}), so we draw no 
conclusion from it. We adopt the post-projection variant as the 
default so that all three losses act on the same embedding space.

\section{Hyperparameter Selection Protocol}
\begin{table}[h!]
\caption{Ablation study on the balancing coefficients of the three loss components used during post-training. Setting the weight of the primary CLIP loss as the baseline ($\lambda_\text{clip}=1.0$), we investigate the impact of varying the proportions of the MSE loss ($\lambda_\text{mse}$) and RKD loss ($\lambda_\text{rkd}$) coefficients. All experiments are conducted on the CC3M dataset based on the pre-trained OpenAI CLIP ViT-B/16 architecture. The best results in each column are highlighted in \textbf{bold}; MMVP (single runs except where a seed standard deviation is given) is within seed-to-seed variance and is not highlighted.}
\label{tab::loss_weights_ablation}
\centering
\resizebox{\textwidth}{!}{
\begin{tabular}{ccc|c|cc|cc|c|c}
\toprule
\multicolumn{3}{c|}{Loss Weights} & Zero-shot & \multicolumn{2}{c|}{MS COCO (R@1)} & \multicolumn{2}{c|}{Flickr30K (R@1)} & Linear Probe & MMVP \\
$\lambda_\text{clip}$ & $\lambda_\text{mse}$ & $\lambda_\text{rkd}$ & 12 datasets Avg & I$\rightarrow$T & T$\rightarrow$I & I$\rightarrow$T & T$\rightarrow$I & 5 datasets Avg & Avg \\
\midrule
1.0 & 1.0 & 0.5 & 62.77 & 52.42 & 36.47 & \textbf{78.00} & 63.06 & \textbf{49.75} & 14.81 \\
1.0 & 0.5 & 1.0 & 62.91 & 52.40 & 36.56 & 77.50 & 63.32 & 47.29 & 13.33 \\
1.0 & 0.5 & 0.5 & 63.03 & \textbf{52.62} & \textbf{36.76} & 77.80 & \textbf{63.38} & 48.85 & 17.04 \\
1.0 & 1.0 & 1.0 & \textbf{63.15} & 52.20 & 36.18 & 77.80 & 63.22 & 48.74 & 14.81\sd{2.97} \\
\bottomrule
\end{tabular}
}
\end{table}

To ensure no validation leakage, we determine all hyperparameters 
on the ViT-B/16 + CC3M setting only, using the following protocol: \textbf{Temperature $\tau$}: Selected via the sweep in 
    Table~\ref{tab::logit_scale_ablation}. We chose $\tau = 0.01$ 
    as it gives the best zero-shot accuracy and the best or near-best
    retrieval; MMVP was not used for selection. We then \emph{transferred this value 
    unchanged} to ViT-L/14, COCO Caption, and CC12M experiments, 
    without any additional tuning.
    
We perform an ablation study to investigate the influence of the weighting coefficients ($\lambda_\text{clip}, \lambda_\text{mse}, \lambda_\text{rkd}$) for the three loss components. Taking the CLIP loss as our primary objective ($\lambda_\text{clip} = 1.0$), we systematically vary the scales of $\lambda_\text{mse}$ and $\lambda_\text{rkd}$ to observe their collective impact on model performance. The results are summarized in Table \ref{tab::loss_weights_ablation}. 

Different configurations favor different metrics: 
$\lambda_\text{clip} : \lambda_\text{mse} : \lambda_\text{rkd} = 1.0 : 1.0 : 0.5$ 
gives the best linear-probe accuracy, and $1.0 : 1.0 : 1.0$ the best 
zero-shot accuracy. The variation across ratios is small: within 
$0.4$ points on zero-shot classification and within $0.6$ points on 
retrieval. We adopt $1.0 : 1.0 : 1.0$ as the default for simplicity; 
MMVP was not used for this choice.

\section{Results After 1 or 2 epochs of Post-training}
\label{app:epochs}
\begin{table}[h!]
\caption{Performance evaluation of post-training with ComCLIP on the CC3M dataset. Experiments are based on the pre-trained OpenAI CLIP ViT-B/16 architecture. We compare the results obtained after 1 and 2 epochs of post-training against the OpenAI CLIP baseline. The best results in each column are highlighted in \textbf{bold}; MMVP (single run for 2 epochs, $3$-seed mean for 1 epoch) is within seed-to-seed variance and is not highlighted.}
\label{tab::epoch_ablation}
\centering
\resizebox{\textwidth}{!}{
\begin{tabular}{c|c|c|cc|cc|c|c}
\toprule
Method & epochs & Zero-shot & \multicolumn{2}{c|}{MS COCO (R@1)} & \multicolumn{2}{c|}{Flickr30K (R@1)} & Linear Probe & MMVP \\
& & CLS Avg & I$\rightarrow$T & T$\rightarrow$I & I$\rightarrow$T & T$\rightarrow$I & Avg & Avg \\
\midrule
OpenAI CLIP (Baseline) & - & 61.82 & 48.16 & 31.47 & 74.70 & 57.10 & 44.70 & 12.59 \\
\midrule
ComCLIP & 1 & \textbf{63.15} & \textbf{52.20} & \textbf{36.18} & \textbf{77.80} & 63.22 & 48.74 & 14.81\sd{2.97} \\
ComCLIP & 2 & 62.84 & 52.02 & 36.24 & 77.20 & \textbf{63.44} & \textbf{49.98} & 16.30 \\
\bottomrule
\end{tabular}
}
\end{table}

To investigate the impact of multiple epochs, we train ComCLIP for 1–2 epochs on the ViT-B/16 model using the CC3M dataset. Table~\ref{tab::epoch_ablation} indicates that extending the training duration did not yield a significant improvement in overall performance, so we use a single epoch throughout.

\section{Geometric Metric for Preserving the Pre-trained Image-Text Feature Space}
\label{sec:text_anchor_kl}
\begin{table}[t!]
\centering
\caption{Text-anchored geometric diagnostic of models post-trained using only $\mathcal{L}_{\text{clip}}$ across different temperature ($\tau$) values. A lower Text Anchor KL ($\downarrow$, closer to 0) indicates that the post-trained visual feature manifold has not suffered from severe structural drift relative to the frozen text features.}
\label{tab::text_anchor_kl}
\begin{tabular}{c|c|c}
\toprule
$\tau$ & \textit{logit Scale} ($1/\tau$) & Text Anchor KL ($\downarrow$) \\
\midrule
0.07  & 14.29  & 5.50 \\
0.03  & 33.33  & 2.18 \\
0.02  & 50.00  & 1.25 \\
0.01  & 100.00 & \textbf{0.48} \\
0.007 & 142.86 & 0.56 \\
0.005 & 200.00 & 1.15 \\
\bottomrule
\end{tabular}
\end{table}

We treat the frozen text features as anchor points and compute the KL divergence of the image feature distribution before and after post-training—referred to as Text-Anchored KL Divergence. A lower KL divergence indicates that the manifold of image-text feature in contrastive space has been better preserved.

Table~\ref{tab::text_anchor_kl} demonstrates that when utilizing the temperature ($\tau=0.07$) used for initialization during the CLIP pre-training phase, the KL divergence reaches a value of 5.50; this indicates that, relative to the frozen text anchors, the visual features have undergone severe manifold drift. Conversely, the "small $\tau$" mechanism we adopted effectively suppresses this drift, resulting in a KL divergence of only 0.48. This is consistent with the gradient analysis in Sec.~\ref{sec:method-clip}: with a small $\tau$, already-separated pairs receive vanishing gradient, so the image features move little relative to the frozen text anchors. The KL is not monotone in $\tau$ (it rises again at $\tau = 0.005$), matching the slight decline in zero-shot accuracy at the smallest temperature.

\section{Training Curves of \texorpdfstring{$\mathcal{L}_{\text{clip}}$}{L\_clip} under Different \texorpdfstring{$\tau$}{tau} Settings}
\begin{figure}[t!]
\centering
\includegraphics[width=0.8\linewidth]{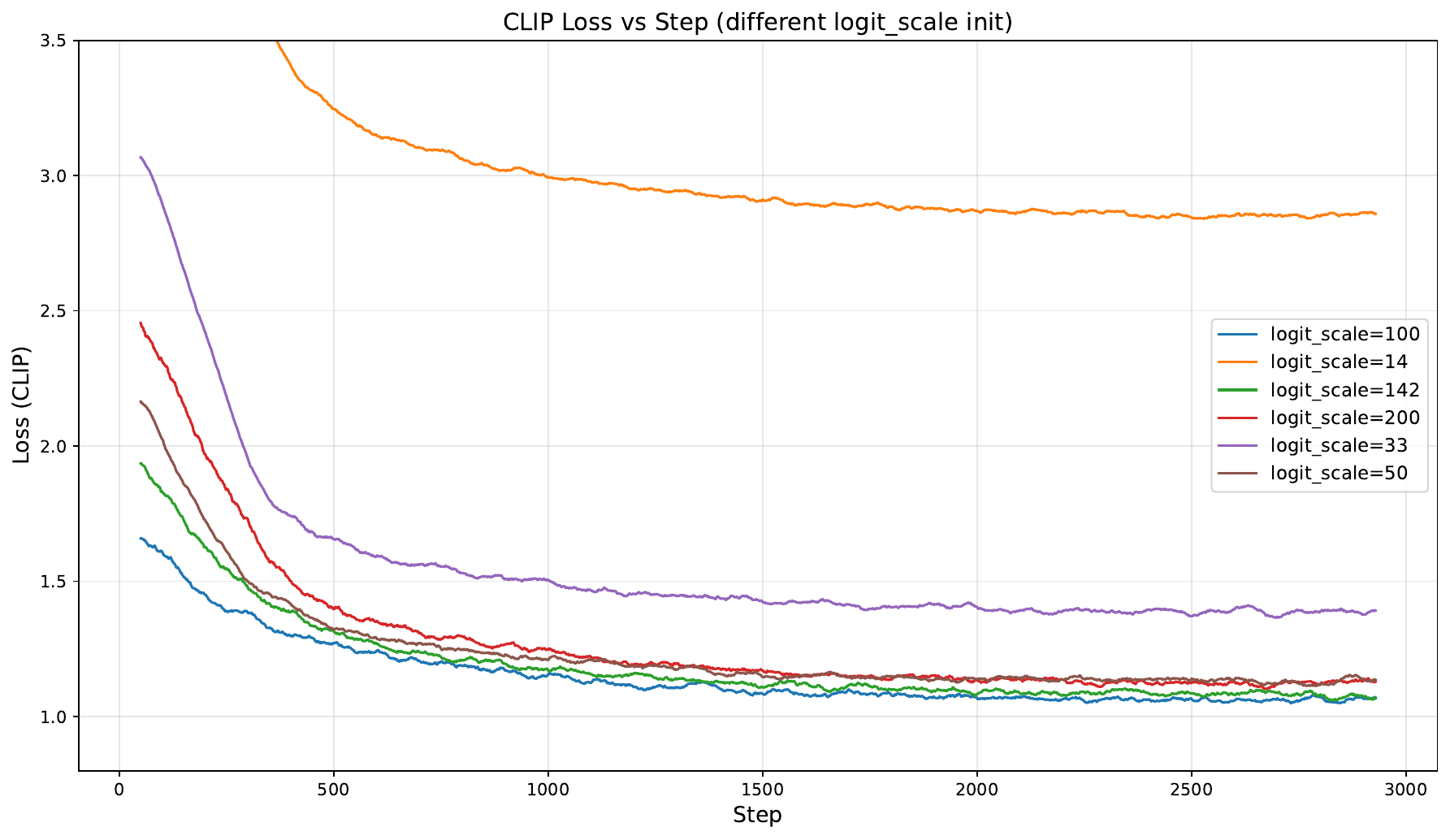}
\caption{\textbf{$\mathcal{L}_{\text{clip}}$ Training Curve.} With the temperature $\tau$ ($\textit{logit\_scale}=1/\tau$) set to various values, post-training was performed using only $\mathcal{L}_{\text{clip}}$. Convergence of the loss function was optimal when $\textit{logit\_scale} = 100$.}
\label{fig:cliploss}
\vspace{-1em}
\end{figure}

Figure~\ref{fig:cliploss} visualizes the variation of $\mathcal{L}_{\text{clip}}$ as a function of training steps under different settings of $\tau$. When using the temperature $\tau=0.07$ (i.e., $\textit{logit\_scale} \approx 14$)—the value initialized by the pre-trained CLIP model—$\mathcal{L}_{\text{clip}}$ remains relatively high and fails to converge to a lower value within 3,000 steps. As $\tau$ is decreased (equivalently, as $\textit{logit\_scale}$ is increased to 33, 50, and 100), $\mathcal{L}_{\text{clip}}$ gradually declines; notably, the convergence curve for $\textit{logit\_scale}=100$ is the most optimal. When $\tau$ is further reduced—corresponding to an increase in $\textit{logit\_scale}$ to 142 and 200—$\mathcal{L}_{\text{clip}}$ increases slightly; nevertheless, it still converges to a favorable level within 3,000 steps, demonstrating that an appropriately small value of $\tau$ is effective for the post-training process.

\section{Detailed Zero-shot Experimental Results}
\begin{table*}[t!]
\centering
\caption{Top-1 zero-shot classification accuracy on 12 public benchmarks. The best and second-best results in each model block are highlighted in \textbf{bold} and \underline{underline}, respectively.}
\vspace{10pt}
\label{tab::zeroshot}
\resizebox{\textwidth}{!}{
\begin{tabular}{c|c|c|c|c|cccccccccccc}
\toprule
Model & Dataset & epochs & Method & Avg & IN-1k & Cifar10 & Cifar100 & CalTech & FER & Pets & DTD & RESISC & EuroSAT & PCAM & IN-S & IN-O \\ \midrule

\multirow{10}{*}{ViT-B/16} & —— & - & OpenAI CLIP & 61.82 & 68.35 & 90.00 & 65.61 & 82.17 & 46.39 & 88.99 & 44.95 & 58.19 & 55.91 & 50.73 & 48.24 & 42.30 \\
 & ImageNet-1k & 2 & KUEA & 62.07 & 68.52 & 90.69 & 67.08 & 82.14 & 46.45 & 89.21 & 45.00 & 58.14 & 55.26 & 51.01 & 48.23 & 43.15 \\
 & CC3M & 2 & KUEA & 61.99 & 68.40 & 90.27 & 66.52 & 82.19 & 46.53 & 89.02 & 45.05 & 58.33 & 56.00 & 51.12 & 48.25 & 42.15 \\
 & COCO Caption & 1 & CLIP-Refine & 62.99 & 67.74 & 90.50 & 68.28 & 83.40 & 49.80 & 89.75 & 44.68 & 59.35 & 56.20 & 55.18 & 48.14 & 42.90 \\
 & CC3M & 1 & CLIP-Refine & 61.59 & 63.30 & 89.56 & 67.25 & 84.29 & 49.47 & 81.47 & 44.63 & 59.97 & 59.46 & 54.23 & 44.28 & 41.20 \\
 & CC3M & 2 & CLIP-Refine & 61.10 & 62.71 & 88.46 & 66.71 & 83.96 & 47.80 & 80.76 & 44.36 & 59.21 & \underline{60.17} & 54.60 & 43.67 & 40.75 \\ 
 & COCO Caption & 1 & ComCLIP & 62.97 & 68.89 & 91.01 & 67.99 & 81.71 & 47.20 & 90.13 & \underline{46.81} & 58.63 & 57.56 & \textbf{55.92} & 48.36 & 41.45 \\
 & CC3M & 1 & ComCLIP & \textbf{63.15} & 69.31 & \textbf{91.50} & \textbf{68.51} & 82.70 & 49.55 & 89.67 & \textbf{46.81} & 59.02 & 55.76 & 53.24 & 48.95 & \underline{42.80} \\
 & CC3M & 2 & ComCLIP & 62.84 & \textbf{69.38} & 91.02 & 67.80 & 82.53 & 48.57 & 89.59 & 46.60 & 58.92 & 56.15 & 52.28 & \textbf{48.97} & 42.25 \\
 & CC12M & 1 & ComCLIP & \underline{63.04} & \underline{69.18} & \underline{90.66} & \underline{66.66} & 81.94 & 47.67 & 89.37 & 46.12 & 59.11 & 58.74 & 54.49 & \underline{49.44} & \textbf{43.10} \\ \midrule

\multirow{8}{*}{ViT-L/14} & —— & - & OpenAI CLIP & 66.12 & 75.54 & 95.19 & 71.05 & 83.27 & 49.96 & 93.24 & 55.27 & 63.33 & 62.65 & 52.01 & 59.59 & 32.35 \\
 & ImageNet-1k & 2 & KUEA & 66.89 & 75.93 & \textbf{95.95} & \textbf{75.13} & 83.78 & 50.06 & 93.24 & 55.69 & 64.08 & 62.30 & 52.15 & 59.93 & \textbf{34.45} \\
 & CC3M & 2 & KUEA & 66.36 & 75.53 & 95.53 & 72.44 & 83.50 & 50.03 & 93.35 & 55.37 & 63.57 & 62.67 & 52.02 & 59.70 & 32.65 \\
 & COCO Caption & 1 & CLIP-Refine & 66.31 & 74.16 & 94.76 & 70.31 & \textbf{84.52} & 48.75 & 91.39 & 55.74 & 63.92 & 66.80 & \textbf{54.40} & 58.68 & 32.25 \\ 
 & COCO Caption & 1 & ComCLIP & 67.09 & 75.82 & 95.18 & 74.07 & 82.96 & 52.72 & 93.35 & 55.90 & 63.56 & 66.74 & 52.79 & 59.95 & 32.05 \\
 & CC3M & 1 & ComCLIP & 67.75 & 76.10 & 95.37 & 73.89 & 82.53 & 53.89 & 93.35 & 56.76 & \textbf{64.62} & \textbf{69.19} & 53.63 & \underline{60.25} & \underline{33.45} \\
 & CC3M & 2 & ComCLIP & \textbf{67.78} & \textbf{76.11} & 95.25 & 73.99 & 82.73 & \textbf{54.07} & \underline{93.46} & \textbf{56.91} & 64.46 & 68.80 & 53.98 & 60.20 & 33.35 \\
 & CC12M & 1 & ComCLIP & \textbf{67.78} & \underline{75.81} & \underline{95.64} & \underline{74.67} & 83.37 & \underline{53.78} & \textbf{93.54} & \underline{55.69} & \underline{64.54} & \underline{67.57} & \underline{54.84} & \textbf{60.43} & \underline{33.45} \\ \midrule

\multirow{3}{*}{\shortstack[l]{ViT-SO\\400M/14}} & —— & - & SigLIP & 71.24 & 82.02 & 96.87 & 84.30 & 86.01 & \textbf{49.89} & 95.26 & 71.06 & 72.56 & 62.74 & 50.55 & \textbf{74.02} & 29.60 \\
 & ImageNet-1k & 2 & KUEA* & \underline{71.67} & \underline{82.18} & \underline{96.88} & \underline{84.89} & \textbf{86.11} & 49.16 & \textbf{95.42} & \underline{71.06} & \textbf{72.84} & \textbf{65.54} & \underline{50.11} & 73.92 & \textbf{31.90} \\
 & CC3M & 1 & ComCLIP & \textbf{72.15} & \textbf{82.50} & \textbf{97.31} & \textbf{85.14} & \underline{85.54} & \underline{49.14} & 95.26 & \textbf{71.91} & \underline{72.48} & \underline{62.11} & \textbf{60.33} & \underline{74.38} & \underline{29.70} \\ 
\bottomrule
\end{tabular}
}
\end{table*}

Table~\ref{tab::zeroshot} presents per-dataset single-run results 
for the baselines and ComCLIP across the $12$ zero-shot 
classification benchmarks. On ViT-B/16, ComCLIP and CLIP-Refine 
(COCO Caption) are at parity in average zero-shot accuracy; over 
three seeds the difference ($63.00_{\pm0.13}$ vs.\ 
$62.87_{\pm0.11}$) is not significant (Table~\ref{tab::seeds}). When 
trained on CC3M, CLIP-Refine drops to $61.59$, notably on ImageNet-1K 
($63.30$ vs.\ $68.35$ for the original CLIP), whereas ComCLIP varies 
by at most $0.2$ across the three post-training datasets.

\section{Detailed Image-text Retrieval Experimental Results}
\begin{table}[t!]
\caption{Zero-shot image-text and text-to-image retrieval results on MSCOCO and Flickr30K benchmarks. The best and second-best results in each model block are highlighted in \textbf{bold} and \underline{underline}, respectively.}
\label{tab::retrieval}
\centering
\resizebox{\textwidth}{!}{
\begin{tabular}{c|c|c|c| ccc ccc | ccc ccc}
\toprule
\multirow{3}{*}{Model} & \multirow{3}{*}{Dataset} & \multirow{3}{*}{epochs} & \multirow{3}{*}{Method} & \multicolumn{6}{c|}{MSCOCO} & \multicolumn{6}{c}{Flickr30K} \\
& & & & \multicolumn{3}{c}{Image-to-Text} & \multicolumn{3}{c|}{Text-to-Image} & \multicolumn{3}{c}{Image-to-Text} & \multicolumn{3}{c}{Text-to-Image} \\
& & & & R@1 & R@5 & R@10 & R@1 & R@5 & R@10 & R@1 & R@5 & R@10 & R@1 & R@5 & R@10 \\
\midrule

\multirow{10}{*}{ViT-B/16}
& —— & - & OpenAI CLIP & 48.16 & 72.58 & 81.88 & 31.47 & 55.62 & 66.73 & 74.70 & 92.90 & 96.40 & 57.10 & 81.36 & 88.20 \\
& ImageNet-1k & 2 & KUEA & 48.48 & 72.98 & 82.24 & 31.99 & 56.35 & 67.21 & 75.00 & 93.10 & 96.60 & 57.62 & 81.82 & 88.66 \\
& CC3M & 2 & KUEA & 48.00 & 72.64 & 81.76 & 31.47 & 55.79 & 66.71 & 75.00 & 92.80 & 96.60 & 57.00 & 81.46 & 88.38 \\
& COCO Caption & 1 & CLIP-Refine & \textbf{53.96} & \textbf{79.02} & \textbf{86.60} & \textbf{37.67} & \textbf{63.83} & \textbf{74.26} & \textbf{80.90} & \textbf{95.40} & \textbf{97.50} & \textbf{64.64} & \textbf{87.34} & \textbf{92.46} \\
& CC3M & 1 & CLIP-Refine & 41.12 & 67.54 & 77.88 & 33.29 & 58.50 & 69.12 & 66.20 & 91.30 & 95.40 & 60.18 & 84.64 & 90.64 \\
& CC3M & 2 & CLIP-Refine & 39.38 & 66.64 & 77.52 & 33.73 & 58.87 & 69.68 & 65.50 & 90.10 & 95.30 & 60.42 & 85.12 & 90.96 \\ 
& COCO Caption & 1	& ComCLIP & \underline{52.24} & \underline{76.56} & \underline{84.60} & 35.36 & 61.09 & 71.78 & \underline{79.50} & 94.70 & 96.70 & 61.36 & 84.62 & 90.80 \\
& CC3M & 1 & ComCLIP & 52.20 & 76.26 & 84.40 & 36.18 & 61.44 & 72.11 & 77.80 & 94.30 & \underline{97.40} & 63.22 & 86.06 & \underline{91.88} \\
& CC3M & 2 & ComCLIP & 52.02 & 75.86 & 84.06 & \underline{36.24} & \underline{61.51} & \underline{72.19} & 77.20 & 94.20 & 96.90 & \underline{63.44} & \underline{86.20} & 91.84 \\
& CC12M & 1 & ComCLIP & 51.58 & 75.84 & 83.78 & 35.56 & 60.70 & 70.99 & 77.10 & \underline{94.80} & \textbf{97.50} & 62.42 & 85.48 & 91.32 \\
\midrule

\multirow{8}{*}{ViT-L/14}
& —— & - & OpenAI CLIP & 50.04 & 73.62 & 82.88 & 34.29 & 58.88 & 69.16 & 77.90 & 94.60 & 98.10 & 59.80 & 83.48 & 89.26 \\
& ImageNet-1k & 2 & KUEA & 50.88 & 75.38 & 83.80 & 35.66 & 60.47 & 70.64 & 79.50 & 95.30 & 98.30 & 61.18 & 84.70 & 90.24 \\
& CC3M & 2 & KUEA & 50.30 & 74.22 & 83.08 & 34.68 & 59.25 & 69.43 & 78.00 & 94.80 & 98.30 & 60.02 & 83.64 & 89.66 \\
& COCO Caption & 1 & CLIP-Refine & 53.28 & 77.42 & 85.30 & 38.21 & 63.05 & 73.24 & 80.10 & 96.20 & 98.30 & 64.56 & 87.86 & \underline{92.74} \\ 
& COCO Caption & 1 & ComCLIP & 54.32 & \textbf{77.96} & 85.78 & 38.79 & 63.93 & 73.77 & 81.80 & \textbf{97.00} & \textbf{99.10} & 64.98 & 86.76 & 92.32 \\
& CC3M & 1 & ComCLIP & \underline{54.56} & \textbf{77.96} & \textbf{85.98} & \underline{39.40} & \underline{64.45} & \underline{74.12} & \textbf{82.60} & \underline{96.30} & 98.80 & \underline{65.82} & \underline{88.22} & 92.62 \\
& CC3M & 2 & ComCLIP & \textbf{54.88} & \underline{77.46} & \underline{85.82} & \textbf{39.63} & \textbf{64.79} & \textbf{74.31} & \underline{82.10} & \underline{96.30} & \underline{98.90} & \textbf{66.28} & \textbf{88.52} & \textbf{92.98} \\
& CC12M	& 1 & ComCLIP & 53.74 & 77.30 & 85.38 & 38.63 & 63.73 & 73.35 & 80.10 & 95.30 & 98.40 & 65.00 & 87.74 & 92.14 \\
\midrule

\multirow{2}{*}{\begin{tabular}{@{}c@{}}ViT-SO \\ 400M/14\end{tabular}} 
& —— & - & SigLIP & 68.16 & 87.46 & 92.86 & 52.20 & 75.71 & 83.37 & 88.90 & 98.50 & 99.60 & 74.80 & 92.24 & 95.22 \\
& CC3M & 1 & ComCLIP & \textbf{70.06} & \textbf{88.24} & \textbf{93.60} & \textbf{53.05} & \textbf{76.48} & \textbf{83.84} & \textbf{89.50} & \textbf{98.90} & \textbf{99.80} & \textbf{75.78} & \textbf{92.88} & \textbf{95.94} \\
\bottomrule
\end{tabular}
}
\end{table}

Table~\ref{tab::retrieval} presents the detailed results (R@1, R@5, and R@10) of the baseline model and our proposed ComCLIP on two image-text retrieval datasets, covering various model architectures, training datasets, and training epochs.

On ViT-B/16, CLIP-Refine (COCO Caption) achieves the best retrieval 
performance on all metrics, and ComCLIP is second. On ViT-L/14, 
ComCLIP achieves higher retrieval than KUEA and CLIP-Refine (COCO 
Caption) in these single runs.

\section{Detailed Linear Probe Experimental Results}
Table~\ref{tab::detailed_linear} presents detailed results for the baselines and our proposed ComCLIP across 5 linear probe tests, covering various model architectures, training datasets, and epochs.

On ViT-B/16, ComCLIP achieves higher average linear-probe accuracy 
than KUEA and CLIP-Refine on all three post-training datasets; the 
advantage over CLIP-Refine is significant over three seeds 
(Table~\ref{tab::seeds}) and holds under dataset-matched comparisons 
(Sec.~\ref{sec:exp-main}). On ViT-L/14, KUEA (ImageNet-1K) achieves 
the best average, plausibly aided by the overlap between its 
post-training data and the ImageNet-1K probe; ComCLIP is second and 
varies by at most $1.3$ across post-training datasets for one epoch.
\begin{table}[h]
\caption{Linear probe evaluation results on various datasets. The best and second-best results in each model block are highlighted in \textbf{bold} and \underline{underline}, respectively.}
\label{tab::detailed_linear}
\centering
\resizebox{\textwidth}{!}{
\begin{tabular}{c|c|c|c|c|ccccc}
\toprule
Model & Dataset & epochs & Method & Avg & IN-1K & SVHN & GTSRB & CLEVR Dist. & CLEVR Counts \\
\midrule
\multirow{10}{*}{ViT-B/16}
& —— & - & OpenAI CLIP & 44.70 & 66.34 & 45.19 & 56.94 & 31.37 & 23.68 \\
& IN-1k & 2 & KUEA & 44.65 & 66.82 & 47.85 & 58.19 & 28.34 & 22.07 \\
& CC3M & 2 & KUEA & 44.72 & 66.46 & 46.22 & 57.48 & 30.83 & 22.62 \\
& COCO Caption & 1 & CLIP-Refine & 42.01 & \textbf{67.51} & 44.53 & 56.60 & 18.58 & 22.83 \\
& CC3M & 1 & CLIP-Refine & 44.89 & 66.22 & 45.61 & 56.94 & 31.38 & 24.28 \\
& CC3M & 2 & CLIP-Refine & 43.52 & 65.74 & 45.20 & 56.76 & 25.89 & 23.99 \\ 
& COCO Caption	& 1	& ComCLIP & 47.23 & 67.00 & 46.27 & 57.13 & 29.49 & 36.25 \\
& CC3M & 1 & ComCLIP & \underline{48.74} & \underline{67.15} & \textbf{48.06} & \textbf{58.26} & \underline{32.69} & \underline{37.53} \\
& CC3M & 2 & ComCLIP & \textbf{49.98} & 66.62 & \underline{46.96} & \underline{57.66} & \textbf{38.11} & \textbf{40.55} \\
& CC12M	& 1	& ComCLIP & 46.56 & 66.69 & 45.86 & 57.09 & 26.38 & 36.79 \\
\midrule

\multirow{8}{*}{ViT-L/14}
& —— & - & OpenAI CLIP & 54.90 & 72.07 & 65.21 & 72.96 & 22.97 & 41.29 \\
& IN-1k & 2 & KUEA & \textbf{59.92} & \textbf{73.37} & \textbf{71.04} & \underline{73.77} & \textbf{28.99} & \textbf{52.42} \\
& CC3M & 2 & KUEA & 55.82 & 72.45 & \underline{66.79} & 73.59 & 23.89 & 42.37 \\
& COCO Caption & 1 & CLIP-Refine & 50.83 & 72.84 & 64.11 & 72.84 & 21.10 & 23.28 \\ 
& COCO Caption & 1	& ComCLIP & \underline{58.28} & 72.11 & 66.71 & \textbf{74.46} & \underline{26.43} & 51.69 \\
& CC3M & 1 & ComCLIP & 56.99 & \underline{72.85} & 64.78 & 73.53 & 21.49 & \underline{52.29} \\
& CC3M & 2 & ComCLIP & 56.60 & 72.63 & 64.65 & 73.20 & 21.03 & 51.47 \\
& CC12M & 1 & ComCLIP & 57.42 & 72.71 & 65.50 & 73.65 & 24.25 & 50.99 \\
\midrule

\multirow{2}{*}{\begin{tabular}{@{}l@{}}ViT-SO \\ 400M/14\end{tabular}} 
& —— & - & SigLIP & 70.87 & 80.78 & \textbf{78.02} & \textbf{89.63} & \textbf{37.94} & 67.98 \\
& CC3M & 1 & ComCLIP & \textbf{70.88} & \textbf{80.84} & \textbf{78.02} & \textbf{89.63} & 37.93 & \textbf{67.99} \\
\bottomrule
\end{tabular}
}
\end{table}

\section{Detailed MMVP Experimental Results}
Table~\ref{tab::detailed_mmvp} presents detailed results for the baselines and our proposed ComCLIP on MMVP benchmark, covering various model architectures, training datasets, and epochs.

MMVP contains $135$ evaluated image pairs in
total and only $15$ pairs per category, so single-run per-category 
values move in steps of $6.67$ points and should not be interpreted 
individually. The figures we report are the multi-seed means in 
Table~\ref{tab::seeds}: on ViT-B/16, ComCLIP ($14.81_{\pm2.97}$) and 
CLIP-Refine ($16.05_{\pm1.14}$) are indistinguishable; on ViT-L/14, 
ComCLIP ($24.20_{\pm0.86}$) is above CLIP-Refine trained on the same 
data ($19.01_{\pm1.86}$). Generative enhancement methods report 
higher MMVP on ViT-L/14 (Appendix~\ref{app:generative}).
\begin{table}[t!]
\caption{Detailed MMVP benchmark results (single runs; each category has only $15$ pairs, so one pair corresponds to $6.67$ points). $^\ddagger$: per-category breakdown of one seed; on ViT-B/16 this seed is the maximum of three (3-seed mean $14.81_{\pm2.97}$), and on ViT-L/14 it is one of three seeds (3-seed mean $24.20_{\pm0.86}$). The multi-seed means in Table~\ref{tab::seeds} are the figures we report; per-category and single-run differences are not used for any conclusion. The best results in each model block are highlighted in \textbf{bold}.}
\label{tab::detailed_mmvp}
\centering
\resizebox{\textwidth}{!}{
\begin{tabular}{c|c|c|c| c|ccccccccc}
\toprule
Model & Dataset & epochs & Method & Avg & 
\begin{tabular}{@{}c@{}}Orientation \\ and\end{tabular} & 
\begin{tabular}{@{}c@{}}Presence of \\ Specific\end{tabular} & 
\begin{tabular}{@{}c@{}}State and \\ Condition\end{tabular} & 
\begin{tabular}{@{}c@{}}Quantity \\ and\end{tabular} & 
\begin{tabular}{@{}c@{}}Positional \\ and\end{tabular} & 
\begin{tabular}{@{}c@{}}Color and \\ Appearance\end{tabular} & 
\begin{tabular}{@{}c@{}}Structural \\ and\end{tabular} & 
Texts & 
\begin{tabular}{@{}c@{}}Viewpoint and \\ Perspective\end{tabular} \\
\midrule

\multirow{10}{*}{ViT-B/16}
& —— & - & OpenAI CLIP & 12.59 & 6.67 & 0.00 & 26.67 & 13.33 & 13.33 & 20.00 & 13.33 & 0.00 & 20.00 \\
& ImageNet-1k & 2 & KUEA & 11.85 & 0.00 & 0.00 & 20.00 & 13.33 & 13.33 & 26.67 & 13.33 & 0.00 & 20.00 \\
& CC3M & 2 & KUEA & 14.07 & 6.67 & 0.00 & 26.67 & 20.00 & 13.33 & 20.00 & 20.00 & 0.00 & 20.00 \\
& COCO Caption & 1 & CLIP-Refine & 17.04 & 13.33 & 6.67 & 20.00 & 0.00 & 20.00 & 46.67 & 33.33 & 0.00 & 13.33 \\
& CC3M & 1 & CLIP-Refine & 19.26 & 13.33 & 0.00 & 26.67 & 6.67 & 26.67 & 53.33 & 33.33 & 0.00 & 13.33 \\
& CC3M & 2 & CLIP-Refine & \textbf{20.74} & 13.33 & 0.00 & 33.33 & 20.00 & 20.00 & 40.00 & 33.33 & 6.67 & 20.00 \\
& COCO Caption & 1 & ComCLIP & 15.56 & 13.33 & 6.67 & 33.33 & 6.67 & 13.33 & 20.00 & 33.33 & 0.00 & 13.33 \\
& CC3M & 1 & ComCLIP$^\ddagger$ & 17.78 & 6.67 & 0.00 & 26.67 & 0.00 & 13.33 & 46.67 & 33.33 & 6.67 & 26.67 \\
& CC3M & 2 & ComCLIP & 16.30 & 6.67 & 0.00 & 33.33 & 0.00 & 13.33 & 46.67 & 33.33 & 6.67 & 6.67 \\
& CC12M & 1 & ComCLIP & 15.56 & 6.67 & 0.00 & 33.33 & 0.00 & 6.67 & 26.67 & 40.00 & 6.67 & 20.00 \\
\midrule

\multirow{8}{*}{ViT-L/14}
& —— & - & OpenAI CLIP & 17.78 & 6.67 & 13.33 & 20.00 & 13.33 & 6.67 & 53.33 & 26.67 & 6.67 & 13.33 \\
& ImageNet-1k & 2 & KUEA & 17.78 & 6.67 & 20.00 & 20.00 & 6.67 & 6.67 & 40.00 & 33.33 & 6.67 & 20.00 \\
& CC3M & 2 & KUEA & 18.52 & 13.33 & 13.33 & 20.00 & 13.33 & 6.67 & 46.67 & 20.00 & 13.33 & 20.00 \\
& COCO Caption & 1 & CLIP-Refine & 17.04 & 6.67 & 13.33 & 26.67 & 13.33 & 6.67 & 33.33 & 13.33 & 13.33 & 26.67 \\
& COCO Caption & 1 & ComCLIP & 20.74 & 6.67 & 13.33 & 33.33 & 13.33 & 6.67 & 53.33 & 13.33 & 13.33 & 33.33 \\
& CC3M & 1 & ComCLIP$^\ddagger$ & 25.19 & 20.00 & 20.00 & 33.33 & 13.33 & 6.67 & 60.00 & 20.00 & 13.33 & 40.00 \\
& CC3M & 2 & ComCLIP & \textbf{25.93} & 20.00 & 26.67 & 40.00 & 13.33 & 6.67 & 40.00 & 26.67 & 13.33 & 46.67 \\
& CC12M & 1 & ComCLIP & 22.22 & 20.00 & 13.33 & 26.67 & 13.33 & 6.67 & 46.67 & 26.67 & 20.00 & 26.67 \\
\midrule

\multirow{2}{*}{\begin{tabular}{@{}c@{}}ViT-SO \\ 400M/14\end{tabular}} 
& —— & - & SigLIP & 39.26 & 26.67 & 20.00 & 53.33 & 40.00 & 20.00 & 66.67 & 40.00 & 26.67 & 60.00 \\
& CC3M & 1 & ComCLIP & \textbf{41.48} & 26.67 & 33.33 & 60.00 & 40.00 & 13.33 & 66.67 & 46.67 & 26.67 & 60.00 \\
\bottomrule
\end{tabular}
}
\end{table}

\section{Comparison with Generative-Model-Based Enhancement}
\label{app:generative}
\begin{table}[t!]
\caption{Comparison with generative-model-based enhancement methods on OpenAI CLIP ViT-L/14 (224px), using the numbers \emph{published} by each paper (base $\rightarrow$ post-trained, as reported in that paper). Evaluation protocols and base values differ across papers (e.g., retrieval R@1 base values), so the rows are not strictly comparable; each row should be read as the change relative to its own base. ``--'': not reported in the form used here. $^\dagger$\,A third-party reproduction from GenHancer's official checkpoints reports ImageNet-1K $40.2$~\citep{un2clip2024}; the two sources disagree. ComCLIP MMVP is the $3$-seed mean (Table~\ref{tab::seeds}).}
\label{tab::generative}
\centering
\resizebox{\textwidth}{!}{
\begin{tabular}{l|l|c|l|c}
\toprule
Method & Enhancement signal & IN-1K zero-shot & Retrieval (as reported) & MMVP-VLM \\
\midrule
DIVA~\citep{diva2024} & diffusion feedback & $75.5 \rightarrow 75.5$ & COCO R@1 I$\rightarrow$T $56.4 \rightarrow 56.7$, T$\rightarrow$I $36.5 \rightarrow 36.6$ & $19.3 \rightarrow 25.9$ \\
GenHancer~\citep{genhancer2024} & generative reconstruction & $75.5 \rightarrow 75.6^\dagger$ & COCO text retrieval R@5 $79.2 \rightarrow 79.4$ & $31.9$ \\
un2CLIP~\citep{un2clip2024} & inverted unCLIP & $75.5 \rightarrow 62.4$ & COCO T$\rightarrow$I R@5 $61.0 \rightarrow 65.5$; Flickr T$\rightarrow$I R@5 $87.3 \rightarrow 90.1$ & $32.6$ \\
\midrule
ComCLIP (ours, CC3M) & contrastive + anchor + RKD, 1 epoch & $75.54 \rightarrow 76.10$ & COCO R@1 I$\rightarrow$T $50.04 \rightarrow 54.56$, T$\rightarrow$I $34.29 \rightarrow 39.40$; Flickr T$\rightarrow$I R@5 $83.48 \rightarrow 88.22$ & $17.78 \rightarrow 24.20_{\pm0.86}$ \\
\bottomrule
\end{tabular}
}
\end{table}

DIVA~\citep{diva2024}, GenHancer~\citep{genhancer2024} and 
un2CLIP~\citep{un2clip2024} improve CLIP's fine-grained perception 
using feedback from a large generative model. A matched retraining 
is outside our compute class, so Table~\ref{tab::generative} reports 
their published numbers on OpenAI CLIP ViT-L/14. Contrary to a common 
characterization (including our own submitted version), these 
methods do not in general sacrifice CLIP's other capabilities: DIVA 
reports unchanged zero-shot classification (27-dataset average 
$69.3 \rightarrow 69.3$; ImageNet-1K $75.5 \rightarrow 75.5$) with 
slightly improved COCO retrieval, and GenHancer reports changes under 
$0.3$ points on ImageNet-1K, SUN397 and COCO retrieval (a third-party 
reproduction from its official checkpoints reports a large ImageNet 
drop, and we cannot resolve the disagreement). un2CLIP reduces 
ImageNet-1K accuracy ($75.5 \rightarrow 62.4$) while improving 
text-to-image retrieval. On MMVP, GenHancer ($31.9$) and un2CLIP 
($32.6$) are well above ComCLIP ($24.20_{\pm0.86}$), and DIVA 
($25.9$) is comparable, although base values differ across papers. 
ComCLIP is therefore not a method for maximizing fine-grained 
perception. It occupies a different point: a single-epoch 
contrastive refinement at a fixed text encoder, without a generative 
model in the loop, whose supported contributions are the temperature 
diagnosis and improved linear-probe transferability.

\section{Broader Impacts}
\label{app:impacts}
ComCLIP is a lightweight post-training method that improves an existing 
vision-language encoder with one epoch on public data, which lowers the 
compute and energy cost of refining CLIP compared with re-training from 
scratch. Because the text encoder is frozen, ComCLIP can be dropped into 
existing CLIP-based systems such as LLaVA without re-training them; more
transferable visual features may benefit downstream applications such as
accessibility tools and visual question answering. On the other hand, ComCLIP inherits the biases of the pretrained 
CLIP and DINOv2 models and of the web-scraped post-training data 
(CC3M, CC12M), and our post-training does not explicitly mitigate them. 
Improved visual encoders could also be misused, for example in 
surveillance applications. We encourage practitioners to evaluate 
downstream systems built on ComCLIP for fairness and safety before 
deployment.

%%%%%%%%%%%%%%%%%%%%%%%%%%%%%%%%%%%%%%%%%%%%%%%%%%%%%%%%%%%%
\newpage
\section*{NeurIPS Paper Checklist}

\begin{enumerate}

\item {\bf Claims}
    \item[] Question: Do the main claims made in the abstract and introduction accurately reflect the paper's contributions and scope?
   \item[] Answer: \answerYes{}%
    \item[] Justification: We summarize the contributions of the paper in the introduction.

\item {\bf Limitations}
    \item[] Question: Does the paper discuss the limitations of the work performed by the authors?
    \item[] Answer: \answerYes{}%
    \item[] Justification: We explicitly discuss the limitations of our method in Section 5 of the paper.

\item {\bf Theory assumptions and proofs}
    \item[] Question: For each theoretical result, does the paper provide the full set of assumptions and a complete (and correct) proof?
    \item[] Answer: \answerNA{}%
    \item[] Justification: There are no theoretical results in this paper.

    \item {\bf Experimental result reproducibility}
    \item[] Question: Does the paper fully disclose all the information needed to reproduce the main experimental results of the paper to the extent that it affects the main claims and/or conclusions of the paper (regardless of whether the code and data are provided or not)?
    \item[] Answer: \answerYes{}%
    \item[] Justification: In Section 4, we have provided the detailed experimental settings for results reproducing.

\item {\bf Open access to data and code}
    \item[] Question: Does the paper provide open access to the data and code, with sufficient instructions to faithfully reproduce the main experimental results, as described in supplemental material?
    \item[] Answer: \answerYes{}%
    \item[] Justification: All post-training and evaluation datasets are public. Code and post-trained checkpoints are released at \url{https://github.com/showstarpro/ComCLIP.git}, and the full experimental setup is given in Section~\ref{sec:exp-setup} and the Appendix.

\item {\bf Experimental setting/details}
    \item[] Question: Does the paper specify all the training and test details (e.g., data splits, hyperparameters, how they were chosen, type of optimizer) necessary to understand the results?
    \item[] Answer: \answerYes{}%
    \item[] Justification: We have provided the detailed experimental settings in Section 4.

\item {\bf Experiment statistical significance}
    \item[] Question: Does the paper report error bars suitably and correctly defined or other appropriate information about the statistical significance of the experiments?
    \item[] Answer: \answerYes{}%
    \item[] Justification: The comparisons that support our main claims are reported as mean $\pm$ standard deviation over seeds (3 seeds, including the ViT-L/14 MMVP and $\mathcal{L}_\text{rkd}$ controls) with two-sided Welch $t$-tests (Table~\ref{tab::seeds}, Table~\ref{tab::rkd_seeds}). Remaining ablations are single runs because of compute cost; they are labelled as such, and we draw no conclusions from differences within seed variance (Section~\ref{sec:exp-ablation-section}).

\item {\bf Experiments compute resources}
    \item[] Question: For each experiment, does the paper provide sufficient information on the computer resources (type of compute workers, memory, time of execution) needed to reproduce the experiments?
    \item[] Answer: \answerYes{}%
    \item[] Justification: We describe the hardware setup, software environment, and codebase for our experiments in detail in Section 4.
    
\item {\bf Code of ethics}
    \item[] Question: Does the research conducted in the paper conform, in every respect, with the NeurIPS Code of Ethics \url{https://neurips.cc/public/EthicsGuidelines}?
    \item[] Answer: \answerYes{}%
    \item[] Justification: Our paper is conducted with the NeurIPS Code of Ethics.

\item {\bf Broader impacts}
    \item[] Question: Does the paper discuss both potential positive societal impacts and negative societal impacts of the work performed?
    \item[] Answer: \answerYes{}%
    \item[] Justification: We discuss positive and negative societal impacts in Appendix~\ref{app:impacts}.
    
\item {\bf Safeguards}
    \item[] Question: Does the paper describe safeguards that have been put in place for responsible release of data or models that have a high risk for misuse (e.g., pre-trained language models, image generators, or scraped datasets)?
    \item[] Answer: \answerNA{}%
    \item[] Justification: This paper does not release generative models. And this paper poses no such risks.

\item {\bf Licenses for existing assets}
    \item[] Question: Are the creators or original owners of assets (e.g., code, data, models), used in the paper, properly credited and are the license and terms of use explicitly mentioned and properly respected?
    \item[] Answer: \answerYes{}%
    \item[] Justification: We conduct experiments in this paper using publicly available datasets and codebases, and cite relevant references.

\item {\bf New assets}
    \item[] Question: Are new assets introduced in the paper well documented and is the documentation provided alongside the assets?
   \item[] Answer: \answerYes{}%
    \item[] Justification: We release the ComCLIP code and post-trained checkpoints with documentation (training/evaluation instructions and license) at \url{https://github.com/showstarpro/ComCLIP.git}.

\item {\bf Crowdsourcing and research with human subjects}
    \item[] Question: For crowdsourcing experiments and research with human subjects, does the paper include the full text of instructions given to participants and screenshots, if applicable, as well as details about compensation (if any)? 
    \item[] Answer: \answerNA{}%
    \item[] Justification: This paper does not involve crowdsourcing nor research with human subjects.

\item {\bf Institutional review board (IRB) approvals or equivalent for research with human subjects}
    \item[] Question: Does the paper describe potential risks incurred by study participants, whether such risks were disclosed to the subjects, and whether Institutional Review Board (IRB) approvals (or an equivalent approval/review based on the requirements of your country or institution) were obtained?
    \item[] Answer: \answerNA{}%
    \item[] Justification: This paper does not involve crowdsourcing nor research with human subjects.

\item {\bf Declaration of LLM usage}
    \item[] Question: Does the paper describe the usage of LLMs if it is an important, original, or non-standard component of the core methods in this research? Note that if the LLM is used only for writing, editing, or formatting purposes and does \emph{not} impact the core methodology, scientific rigor, or originality of the research, declaration is not required.
    \item[] Answer: \answerNA{}%
    \item[] Justification: The core method development in this research does not involve LLMs as any important, original, or non-standard components.

\end{enumerate}

\end{document}